\documentclass[10pt,journal,compsoc]{IEEEtran}
\newif\ifpeerreview

\peerreviewfalse

\usepackage[nocompress]{cite}
\usepackage{url}
\usepackage{amsmath,amssymb,graphicx}
\usepackage{booktabs}
\usepackage{multirow}
\usepackage{lipsum} 
\usepackage{xcolor}
\usepackage{hyperref}
\usepackage[switch]{lineno}

\newcommand{\paperID}{17}

\title{ShuffleFlow: Scalable Posterior Inference for Bayesian Inverse Imaging}

\author{
    Tianao~Li,~\IEEEmembership{Student Member,~IEEE,}
    Tjitske~Starkenburg,
    Yu~Sun,~\IEEEmembership{Member,~IEEE,}
    and~Emma~Alexander,~\IEEEmembership{Member,~IEEE}
    \IEEEcompsocitemizethanks{
        \IEEEcompsocthanksitem T. Li and E. Alexander are with the Department of Computer Science, Northwestern University, Evanston, IL 60208. 
        T. Starkenburg is with the Center for Interdisciplinary Exploration and Research in Astrophysics and the Department of Physics and Astronomy, Northwestern University, Evanston, IL 60208.
        T. Li, E. Alexander, and T. Starkenburg are also with the NSF-Simons AI Institute for the Sky (SkAI), Chicago, IL 60611.
        Y. Sun is with the Department of Electrical and Computer Engineering, Johns Hopkins University, Baltimore, MD 21218.
        \protect\\
        E-mail: ealexander@northwestern.edu
        \IEEEcompsocthanksitem Project website:\href{https://nubivlab.github.io/ShuffleFlow/}{https://nubivlab.github.io/ShuffleFlow/}.
    }
}

\begin{document}

\IEEEtitleabstractindextext{%
\begin{abstract}
Variational inference (VI) is a powerful method for principled posterior inference for scientific inverse imaging.
VI learns the posterior distribution, often with a flow-based network, which can cheaply generate posterior samples upon optimization, and can flexibly incorporate score-based or classic priors.
However, its application to large-scale image reconstruction is severely hindered by the poor scalability of the flow-based networks. 
In this work, we introduce {\em ShuffleFlow}, a scalable VI framework to address this challenge. 
Our method breaks down the problem into three parts: a pixel-unshuffling-based image coordinate sampler, a neural field as feature encoder, and a conditional normalizing flow (CNF) as posterior estimator.
Specifically, our framework partitions an image into a stack of sub-images with pixel-unshuffling and uses a shared CNF to model the joint distribution of the sub-image stack.
We condition the CNF on the output of a neural field, which embeds feature vectors corresponding to pixel-unshuffling sample locations to capture spatial structures, and share the flow's latent variable across the channels to model their correlations.
We demonstrate our method's effectiveness and efficiency on both linear and nonlinear imaging inverse problems, and show its ability to more rapidly generate a high-sample-count posterior than diffusion samplers.
\end{abstract}

}

\maketitle

\IEEEraisesectionheading{
  \section{Introduction}\label{sec:introduction}
}
%
%
%
%

\IEEEPARstart{I}{nverse} problems fundamentally exist in computational imaging systems, which aim to recover latent images $\mathbf{x}$ encoded in noisy and incomplete measurements $\mathbf{y=A(x)+n}$ with physical knowledge of the forward model $\mathbf{A}(\cdot)$. 
This is usually accomplished by solving a maximum-a-posteriori (MAP) problem from a Bayesian perspective, for estimates
\begin{equation}
    \begin{aligned}
        \mathbf{x}^* &= \mathop{\mathrm{arg\,min}}_{\mathbf{x}} \left\{ -\log p(\mathbf{y}|\mathbf{x}) - \log p(\mathbf{x}) \right\},
    \end{aligned}
    \label{eq:map}
\end{equation}
where $p(\mathbf{y}|\mathbf{x})$ is the data likelihood and $p(\mathbf{x})$ is the prior.
These inverse problems are generally ill-posed due to information loss and measurement noise, so reconstructing high-quality images typically requires image priors. For scientific and medical imaging, there is an increasing demand for uncertainty quantification, such as by generating samples to approximate the posterior.

Though diffusion models have recently emerged as a powerful generative model to learn rich prior distributions, which can be used to sample from the Bayesian posterior by conditioning on measurements~\cite{song2022solving,jalal2021robust,kawar2022denoising,graikos2022diffusion,chung2023diffusion,mardani2024a,wu2024principled,zhang2025improving}, diffusion samplers come with several drawbacks.
First, training diffusion models requires large datasets, making them unusable for novel applications where high-quality data are scarce.
Second, drawing each posterior sample requires iterative and repeated diffusion network evaluations, leading to a high cost for generating \textit{high-resolution}, i.e., large sample-count, posteriors. 

Variational inference (VI) provides an alternative to the limitations listed above.
Importantly, after learning the variational distribution, one can efficiently draw samples from the variational network.
This is particularly important when a \textit{high-resolution} posterior is needed for downstream science (e.g., imaging for astronomy~\cite{adam2022posterior,karchev2022strong,feng2024event,barco2025blind}). 
VI methods can also incorporate either classic or modern (e.g., diffusion-based) priors, making them flexible for both data-rich and data-scarce applications. 
Finally, VI has been described as a more principled posterior estimation~\cite{sun2021deep,feng2023score,feng2024variational}, because its only approximation is the flow-network, compared to the approximations in diffusion samplers due to the intractability of the likelihood score~\cite{song2022solving,jalal2021robust,kawar2022denoising,graikos2022diffusion,chung2023diffusion,mardani2024a}.

Despite the aforementioned advantages, VI has seen limited adoption in inverse imaging (though, see~\cite{sun2021deep,sun2022alpha,feng2023score,hong2023robustness,feng2024variational,feng2024event}). This is likely due to challenges in scalability: modeling high-dimensional distributions like image distributions usually requires a large normalizing flow network \cite{sun2021deep,feng2023score}, which leads to high memory and time budgets during optimization that scale poorly with image dimension ($O(N^4)$ for $N \times N$ pixels).
To adapt VI to inverse problems on larger images, prior work either adopts a naive patching~\cite{meng2020gaussianization} or pixelating strategy~\cite{shen2021stochastic,shen2022conditional}, at the cost of decreased image quality and posterior expressiveness.
In this work, our goal is to improve the scalability of VI without the loss of posterior accuracy.

\begin{figure*}[t]
    \centering
    \includegraphics[width=\linewidth]{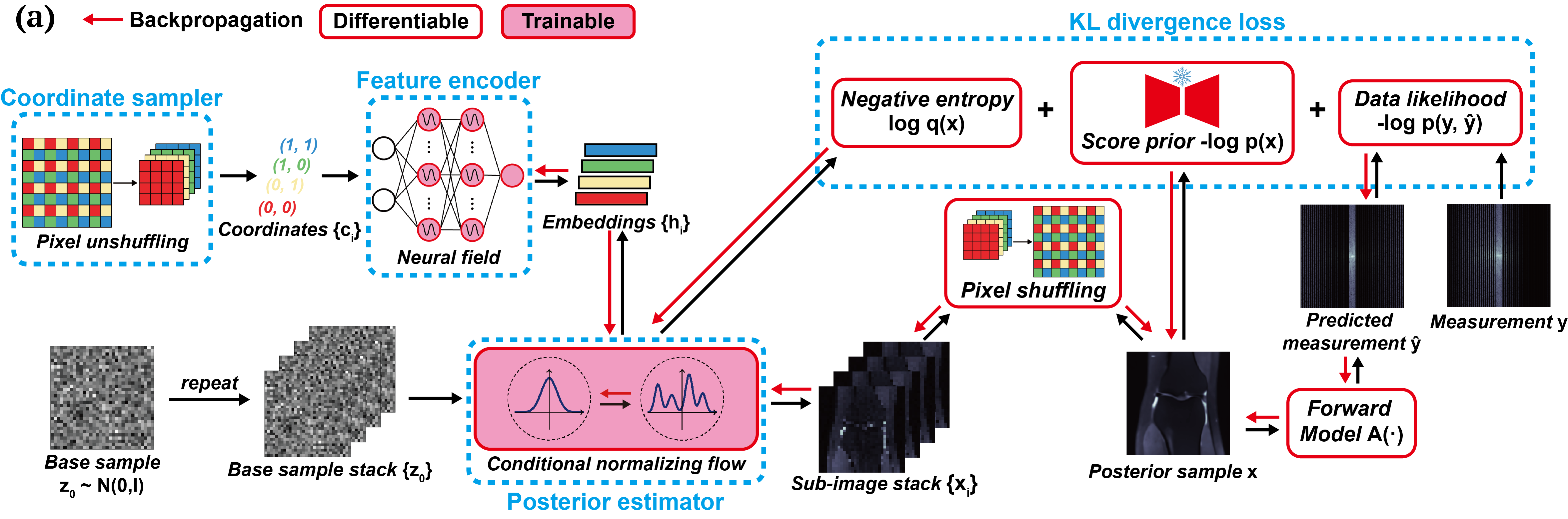}
    \includegraphics[width=0.99\linewidth]{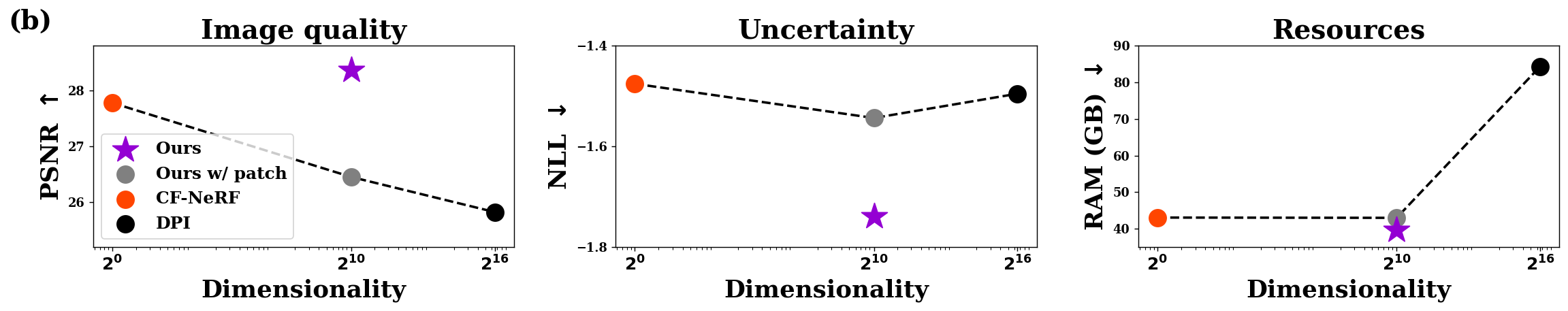}
    \caption{
        {\bf Overview of ShuffleFlow.}
        (a) Overall pipeline of ShuffleFlow.
        The full $N\times N$ image is partitioned into $S^2$ sub-images with shape $\frac{N}{S}\times\frac{N}{S}$ using pixel-unshuffling with a pre-defined downsample factor $S$, and feature embeddings $\{\mathbf{h}_i\}_{i=1,\dots,S^2}$ are calculated from the sampling offset positions using a coordinate-based MLP (NF encoder), implemented with SIREN~\cite{sitzmann2020implicit}.
        We then draw a base sample $\mathbf{z}_0$ with dimension of $\frac{N^2}{S^2}$ from a standard normal distribution and feed it into a conditional normalizing flow $S^2$ times (each time conditioned on an embedding $\mathbf{h}_i$ for a different sub-image $\mathbf{x}_i$) to obtain the pixel-unshuffled image stack $\{\mathbf{x}_i\}_{i=1,\dots,S^2}$.
        We then convert this stack of sub-images into a full image sample using pixel-shuffling, which will then be fed into the VI training loss calculated from the KL divergence, consisting of three terms: data likelihood, prior (score network weights are frozen during optimization), and negative entropy.
        The CNF and NF are trained together at test time on a single measurement (see red arrows for backpropagation). Other priors (e.g., TV) can also be incorporated in place of the score prior.
        (b) 
        For a 256$\times$256 image, previous VI methods either exhaustively model the full image posterior~\cite{sun2021deep} (dimensionality=$2^{16}$) or model an oversimplified one-dimensional posterior~\cite{shen2022conditional} (dimensionality=1). Our method considers 64 subsampled images of 32$\times$32 pixels each (dimensionality=2$^{10}$), which outperforms both existing methods and a naive patching strategy~\cite{meng2020gaussianization}.
        The x-axis shows the posterior dimensionality modeled by each method.
    }
    \label{fig:teaser}
\end{figure*}

We introduce ShuffleFlow, a novel modular VI framework designed for scalable posterior inference in Bayesian inverse imaging.
Inspired by the inherent low-dimensionality of natural images~\cite{carlsson2008local,fefferman2016testing} and the pixel-unshuffling technique that breaks down images to be processed by cheaper networks~\cite{shi2016real,sajjadi2018frame,gu2019self}, our intuition is to break down the full posterior modeling task into several components.
Specifically, ShuffleFlow decomposes the full image into a stack of downsampled sub-images with pixel-unshuffling, and models the joint distribution of the sub-image stack with a shared conditional normalizing flow (CNF) that is significantly lower-dimensional, hence much cheaper. Additionally, we reduce the CNF size by offloading some of the image reconstruction tasks into a neural field (NF) encoder. Specifically, the sampling offset coordinate of the downsampled images is fed into the NF to calculate coordinate-based feature embeddings, which are used as conditional inputs for the CNF.
We show the success of our method using both classic total-variation (TV) prior and a learned score prior to demonstrate its flexibility across data-poor and data-rich settings.
Our contributions are:

1. {\bf Problem decomposition}: We propose a new VI framework for inverse problems that decomposes the posterior modeling of a large image into three parts (see Fig.~\ref{fig:teaser}a): coordinate sampling, partial image representation, and light-weight uncertainty quantification. Specifically, a pixel-unshuffling-based coordinate sampler partitions an image into a stack of sub-images, a neural field encodes spatial features for each sub-image, and a conditional normalizing flow estimates the joint distribution of the sub-image stack.
Our framework's modular structure is agnostic to specific network architectures and allows easy integration of advances in neural fields, normalizing flow networks, and generative priors.

2. {\bf Image quality and uncertainty quantification}:
ShuffleFlow learns complex distributions and achieves SOTA image reconstruction and uncertainty quantification among VI methods (see Fig.~\ref{fig:teaser}b). Across diverse applications, we outperform previous VI methods that fully rely on normalizing flows~\cite{sun2021deep,feng2023score}, or decompose the posterior with naive patching~\cite{meng2020gaussianization} or pixelating~\cite{shen2021stochastic,shen2022conditional} strategies.

3. {\bf Efficient inference for high-sample-count posteriors}: 
ShuffleFlow is a time- and memory-efficient method for posterior estimation compared to existing methods in scientific imaging.
Fig.~\ref{fig:scale} shows a 31-98\% decrease in computational resources for VI, while Fig.~\ref{fig:fpr} demonstrates that we can recover a high-resolution bimodal posterior of 10,000 samples within 16 min, at which point diffusion samplers do not yet show clear bimodal distributions.

In our experiments, our method shows SOTA image reconstruction quality in less time and memory than previous VI methods.
Compared to diffusion samplers, our method is able to quickly generate a large number of posterior samples to discover a bimodal posterior in nonlinear Fourier phase retrieval, which is vital for applications where a high-resolution posterior is needed for downstream science.

\section{Related Work}
\subsection{Image Priors}
Classic methods adopt hand-crafted regularization (e.g., total variation~\cite{rudin1992nonlinear}, sparsity~\cite{candes2007sparsity}) as prior, and solve Eq.~\ref{eq:map} as a regularized maximum likelihood (RML) problem. 
Deep neural networks have been used to learn a prior from data and are incorporated into inverse problem solvers as plugin denoisers~\cite{venkatakrishnan2013plug,chan2016plug,romano2017little,monakhova2019learned,sun2021scalable,li2023galaxy}.
Diffusion models~\cite{song2019generative,ho2020denoising,song2020denoising,song2020score} have emerged as powerful generative models that can capture rich image priors and allow conditional sampling from an approximated posterior, either using projection onto the measurement space~\cite{song2022solving,chung2022score,chung2022improving} or taking gradient steps towards higher data likelihood, such as Score-ALD~\cite{jalal2021robust}, DPS~\cite{chung2023diffusion}, and many others~\cite{kawar2022denoising,graikos2022diffusion,chung2023diffusion}.
RED-Diff~\cite{mardani2024a} used a variational diffusion sampler that uses the diffusion process as regularization, but is mode-seeking and lacks posterior diversity.
Recent diffusion sampling methods avoid the intractability of the likelihood score calculation via split Gibbs sampling (PnP-DM)~\cite{wu2024principled} or decoupled noise annealing (DAPS)~\cite{zhang2025improving}.

\subsection{Neural Fields}

Neural Fields (NFs)~\cite{xie2022neural}, also known as Implicit Neural Representations (INRs), have recently emerged as powerful representations for novel view synthesis~\cite{martin2021nerf,mildenhall2021nerf,barron2021mip} and self-supervised image reconstruction. 
By representing images or physical fields using a coordinate-based Multilayer Perceptron (MLP), NFs generally use a relatively small number of weights (usually orders of magnitude smaller than the signal dimension), making large-scale inverse problems better-posed but also significantly reducing the memory requirements. 
NFs optimize their weights directly on the measurements at test time, bypassing the need for extensive training data, and making them well-suited for novel and complex systems where high-quality labeled datasets are scarce.
The MLP's continuous representation imposes an implicit prior and has been shown to be an effective regularizer for natural signals~\cite{sitzmann2020implicit}. 
Given these advantages, NFs have been successfully applied to a variety of scientific imaging problems~\cite{sun2021coil,zang2021intratomo,liu2022recovery,shen2022nerp,molaei2023implicit,zhou2023fourier,xu2023nesvor,cao2024neural,li2025coordinate}.

\subsection{Normalizing Flows}

Normalizing flows~\cite{rezende2015variational,papamakarios2021normalizing} have been widely used to model complex posterior distributions by transforming a simple base distribution into a desired distribution with invertible functions. 
The invertibility of flow functions enables efficient likelihood calculation through the change-of-variable formula. 
This capability, coupled with their ability to learn intricate distributions, has led to application in various domains, including 3D point cloud generation~\cite{yang2019pointflow, pumarola2020c}, inverse problems in imaging~\cite{sun2021deep,sun2022alpha,feng2023score,feng2024event,feng2024variational,hong2023robustness}, and uncertainty quantification in NeRF~\cite{shen2021stochastic,shen2022conditional}.
In computational imaging, Deep Probabilistic Imaging (DPI)~\cite{sun2021deep,sun2022alpha} is a leading posterior sampling method that leverages a flow-based generative network and variational inference to approximate the posterior and generate posterior samples with hand-crafted regularization.
\cite{feng2023score} incorporates score-based diffusion models as a principled prior into VI frameworks, and~\cite{feng2024variational} proposed a surrogate score-based prior that significantly reduces the associated computation.
However, the computational expense of the flow networks used in these works still scales poorly with the image size, limiting their practical application. 

Previous work has, like the proposed method, attempted to address the scalability problem, but with suboptimal trade-offs.
\cite{meng2020gaussianization} crops an image into several non-correlated patches and fails to capture long-range correlations.
\cite{dai2021sliced} designed a hierarchical architecture, but it still requires at least one full-dimensional flow layer.
CF-NeRF~\cite{shen2022conditional} combines a neural field with a CNF to model complex scenes, but constraints the posterior on a one-dimensional manifold, significantly limiting the expressiveness of the posterior. Of these, CF-NeRF is the method best suited for imaging problems, but its naive single-pixel approach fails to capture posterior complexity as seen in nonlinear inverse problems. 

Note that the choice of downsampling strategies is limited by the invertibility requirement of the normalizing flows, and most common strategies (e.g., average/max pooling, bilinear interpolation) only produce a single sub-image and discard information needed for invertibility.

\subsection{Other Sampling and UQ methods}

Quantifying the uncertainty in reconstructed images is crucial for computational imaging with scientific and medical applications. 
Traditional uncertainty quantification (UQ) methods include MCMC sampling~\cite{bardsley2012mcmc,laumont2022bayesian,cardoso2023monte, sun2024provable, wu2024principled}, Bayesian hypothesis testing~\cite{repetti2019scalable}, and variational inference~\cite{blei2017variational}.
With the recent flourishing of deep learning, Bayesian learning~\cite{neal2012bayesian,gal2016dropout,kendall2017uncertainties} has been applied to quantify the uncertainty of learning-based image reconstruction~\cite{xue2019reliable}. 
\cite{vasconcelos2022uncertainr} provides a Bayesian learning reformulation of NFs, but does not attempt to learn the posterior distribution directly through VI.
Other methods include deep ensembles~\cite{lakshminarayanan2017simple}, equivariant bootstrapping~\cite{pereyra2024equivariant}, and conformal methods~\cite{angelopoulos2022image,ye2025learned}.

\section{Proposed Method}
\label{sec:method}
Our key insight is to break down the problem of image posterior inference into three parts: coordinate sampling, feature encoding, and posterior estimation (Fig.~\ref{fig:teaser}a). 

\subsection{Image Posterior Representation}
We consider an image $\mathbf{x} \in \mathbb{R}^{N^2}$ with size $N\times N$.
Conventional methods~\cite{sun2021deep} represent the probability distribution of this image $q(\mathbf{x})$ using a flow network of dimension $N^2$ with $O(N^4)$ parameters~\cite{dinh2016density}, which scales poorly with the image size and poses challenges to computational resources.
We reduce this significant expense by leveraging the inherent reduction of dimensionality from image pixels to underlying intrinsic dimensions~\cite{carlsson2008local,gong2019intrinsic,fefferman2016testing,pope2021intrinsic,li2025understanding}, which we do by subsampling the image and using a learned feature encoding for each pixel group.

\subsubsection{Coordinate sampling}
In contrast to previous work which either decompose the image into independent patches~\cite{meng2020gaussianization} or correlated pixels~\cite{shen2022conditional}, we decompose the full image into $S^2$ correlated sub-images $\{\mathbf{x}_i\}_{i=1,\dots,S^2}$ with shape $\frac{N}{S}\times\frac{N}{S}$ ($S$ is a divisor of $N$) using pixel-unshuffling~\cite{shi2016real}, such that
$ \{\mathbf{x}_i\}_{i=1,\dots,S^2} = \mathcal{PU}(\mathbf{x};S)$ and $\mathbf{x} = \mathcal{PS}\left(\{\mathbf{x}_i\}_{i=1,\dots,S^2}\right)$, 
where $\mathcal{PU}(\cdot;S)$ is the pixel-unshuffling operation with downsampling factor $S$, and $\mathcal{PS}(\cdot)$ is its inverse, the pixel-shuffling operation.
Note that every sub-image $\mathbf{x}_i$ is assigned a 2D coordinate $\mathbf{c}_i$ based on the offset of the sampling location (see cartoon illustration on the top left of Fig.~\ref{fig:teaser}a).
We refer to this as the pixel-unshuffling {\em coordinate sampling} strategy in this paper.

\subsubsection{Scalable posterior estimation}
Since these sub-images $\{\mathbf{x}_i\}_{i=1,\dots,S^2}$ capture the same coarse global structure of the original image, we use a {\em shared} conditional normalizing flow $\text{CNF}_\theta(\cdot)$ with parameters $\theta$ to model their distributions, with the coordinate information as the condition.
Such low-resolution similarity has also been leveraged in self-supervised image denoising~\cite{huang2021neighbor2neighbor,mansour2023zero}.
This significantly lowers the size of the flow network from $N^2$ to  $\frac{N^2}{S^2}$.
We observe an optimal downsampling factor S=8, leading to a 64$\times$ reduction in dimensionality (see Sec.~\ref{supp:ablations}).
Adopting De Finetti’s Theorem~\cite{cifarelli1996finetti}, we share the base variable $\mathbf{z}_0$ of the CNF across all downsampled images, such that their joint distribution $q \left( \{\mathbf{x}_i\}_{i=1,\dots,S^2} \right)$, which is essentially the distribution of the full image $q(\mathbf{x})$, can be written as a fully-factorized distribution conditioned on this base variable $\mathbf{z}_0$:
\begin{equation}
    q(\mathbf{x}) = q \left( \{\mathbf{x}_i\}_{i=1,\dots,S^2} \right) = \left( \prod_{i=1}^{S^2} q(\mathbf{x}_i|\mathbf{z}_0;\mathbf{c}_i) \right)q(\mathbf{z}_0),
    \label{eq:definetti}
\end{equation}
where $q(\mathbf{x}_i|\mathbf{z}_0;\mathbf{c}_i)$ is the conditional distribution of the CNF, and $q(\mathbf{z}_0)$ is the base variable distribution, for which we use a standard normal distribution.
Sharing base variables lowers the dimensionality of the image distribution from $N^2$ to $\frac{N^2}{S^2}$, which is well aligned with the inherent lower-dimensionality of the image $\mathbf{x}$ and makes learning complex posteriors easier.
It also correlates the downsampled images to prevent noise in image samples originating from pixel independencies~\cite{shen2021stochastic,shen2022conditional}.

\subsubsection{Feature encoding}
Instead of directly using the sampling offset coordinates as conditional inputs for the CNF, we use a neural field $\text{NF}_\phi$ with parameters $\phi$ as a feature encoder to calculate an embedding $\mathbf{h}_i = \mathrm{NF}_\phi(\mathbf{c}_i)$ as the conditional input~\cite{shen2022conditional} for each sub-image $\mathbf{x}_i$. This technique relieves some of the image reconstruction burden from the expensive flow network.

The representation of the full image distribution $q_{\theta,\phi}(\mathbf{x})$ of ShuffleFlow can be written as
\begin{equation}
    \begin{aligned}
        \mathbf{x} \sim q_{\theta,\phi}(\mathbf{x}) \Leftrightarrow \mathbf{x} &= \mathcal{PS} \left( \left\{ \mathrm{CNF}_\theta \left(\mathbf{z}_0,\mathbf{h}_i \right) \right\}_{i=1,\dots,S^2} \right), \\ \mathbf{z}_0 &\sim \mathcal{N}\left(0,\mathbf{I}_{N^2/S^2}\right),
    \end{aligned}
    \label{eq:sampling}
\end{equation}
where $\mathbf{I}_{N^2/S^2}$ is the $\frac{N^2}{S^2}$-dimensional identity matrix.
We observe that using a smaller neural field, using positional encoding, and directly conditioning on the 2D coordinates lead to decreased image quality (see ablations in Sec.~\ref{supp:ablations}).

\begin{figure*}[t]
    \centering
    \includegraphics[width=\linewidth]{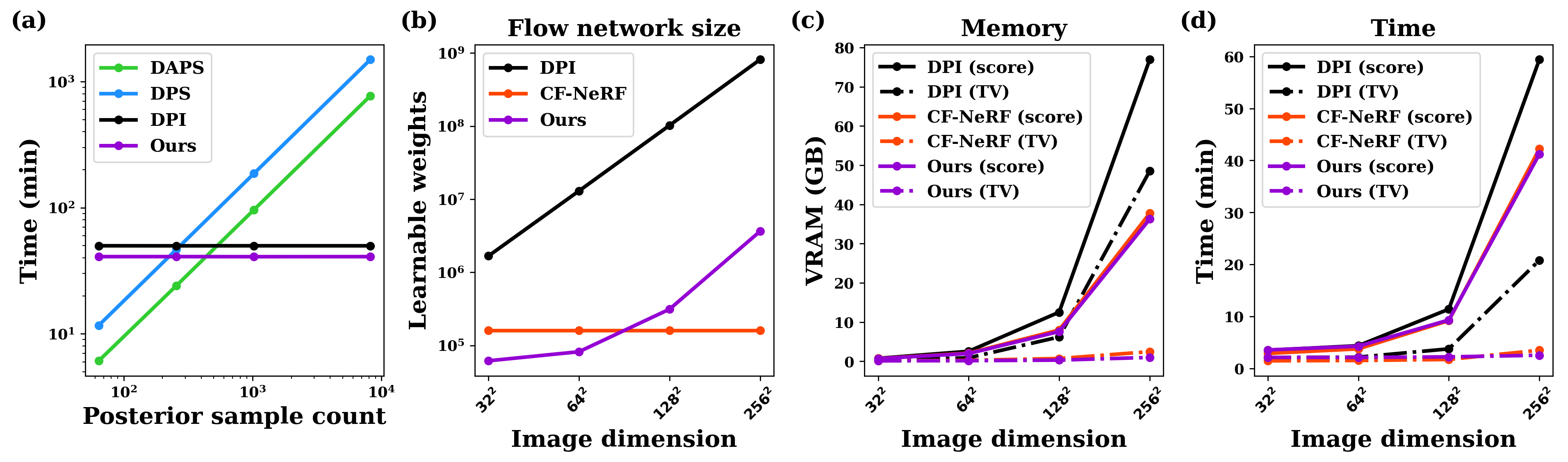}
    \caption{
        {\bf Scalability comparison.} 
        (a) VI methods scale dramatically better than diffusion samplers for high-resolution posterior estimation. 
        (b,c,d) Flow network size, optimization time, and VRAM vs. image dimension across VI methods.
        When the computation is dominated by the flow network (TV prior, dashed lines), our method (purple) shows the theoretically predicted growth trend compared to DPI (black). 
        Even when an expensive score prior dominates (solid lines), our method still shows a substantial efficiency advantage. 
        Experiment settings follow Sec.~\ref{sec:linear} and Tabs.~\ref{tab:linear} \& \ref{supp:inverse_problems}.
    }
    \label{fig:scale}
\end{figure*}

\subsubsection{Efficiency analysis}
In scientific applications, a large number of samples can be needed to characterize the posterior distribution (e.g.,~\cite{adam2022posterior,karchev2022strong,feng2024event,barco2025blind}).
Since the sampling cost grows linearly with the number of samples for diffusion samplers, VI methods are a better solution because they can cheaply generate samples after training (see Fig.~\ref{fig:scale}a). 
Among VI methods, we show theoretically and empirically that our problem decomposition scales nearly as efficiently as pixel-wise inference (see Fig.~\ref{fig:scale}b-d).

While latent-space diffusion models and distillation might reduce the score network size or sampling steps, they do not change the linear scaling curve of diffusion samplers, i.e., the time for generating a high-resolution posterior still grows linearly with sample count.
For methods using Langevin dynamics (e.g., PnP-DM~\cite{wu2024principled}, DAPS~\cite{zhang2025improving}), which require differentiating through the decoder repeatedly, using a latent-space diffusion model might instead increase the sampling cost (see Appx.~B in~\cite{zhang2025improving}).

Mathematically, because the size of the flow networks is proportional to the square of its dimension~\cite{dinh2016density}, we lower the size of the CNF network from $\mathcal{O}(N^4)$ (with dimension $N^2$) to $\mathcal{O}\left(\frac{N^4}{S^4}\right)$ (with dimension $\frac{N^2}{S^2}$).
Although the minimal NF size to well represent an image also depends on the image statistics, it is usually upper bounded by the image size $\mathcal{O}(N^2)$, corresponding to a pixel grid representation~\cite{kimgrids,essakinewe}.
Therefore, the total model size of ShuffleFlow is 
\begin{equation}
    S(N,S) = \mathcal{O}\left(\frac{N^4}{S^4}+N^2\right),
    \label{eq:model_size}
\end{equation}
which will be dominated by the CNF at a large image size.
In Fig.~\ref{fig:scale}b, we show the total model size as a function of image size and observe the polynomial growth for DPI and the constant for CF-NeRF, which has one-dimensional flows, hence a constant network size.

Our method also shows better scalability for optimization time and VRAM (shown in Fig.~\ref{fig:scale}c,d.
Note that the computation is dominated by the flow network when a cheap classic prior is used (dashed line), and we see the theoretically expected trends.
When the score prior is present (solid line), our method still shows better scalability than DPI~\cite{sun2021deep}.

In practice, we observe an optimal image reconstruction and uncertainty calibration at $S=8$ for 256$\times$256 images, leading to a 4096$\times$ reduction in flow network size (see ablations in Sec.~\ref{supp:ablations}). 

For image size $N$, we empirically observe that a downsampling factor $S=\sqrt{N}/2$ (see Sec.~\ref{supp:ablations}) preserves performance, leading to an effective scaling of $O(N^2)$ in Eq.~\ref{eq:model_size}.

\subsection{Optimization}

We use a variational Bayesian approach to train ShuffleFlow's CNF and NF together by minimizing the Kullback-Leibler (KL) divergence between the variational distribution $q_{\theta,\phi}(\mathbf{x})$ and the Bayesian image posterior $p(\mathbf{x}|\mathbf{y})$:
\begin{equation}
    \begin{aligned}
        \theta^*, \phi^* &= \mathop{\mathrm{arg\,min}}_{\theta, \phi} D_{\mathrm{KL}} (q_{\theta,\phi}(\mathbf{x})||p(\mathbf{x}|\mathbf{y})) \\
        &= E_{q_{\theta,\phi}(\mathbf{x})} \left[ - \log p(\mathbf{y}|\mathbf{x}) - \log p(\mathbf{x}) + \log q_{\theta, \phi}(\mathbf{x}) \right],
    \end{aligned}
    \label{eq:KL}
\end{equation}
where the three terms are the data likelihood, prior, and negative entropy. 
The first two terms come from the Bayesian posterior in Eq.~\ref{eq:map}, and the negative entropy term encourages diversity and prevents the variational distribution from collapsing to a point estimate~\cite{sun2021deep}.
During optimization, we calculate Eq.~\ref{eq:KL} with a Monte-Carlo approximation and use explicit data-fitting loss $\mathcal{L}(\cdot)$ and prior $\mathcal{R}(\cdot)$:
\begin{equation}
    \begin{aligned}
        \theta^*, \phi^* = \mathop{\mathrm{arg\,min}}_{\theta, \phi} \sum_{j=1}^M &\left\{ \mathcal{L}\left(\mathbf{y}, \mathbf{A}(\mathbf{x}^j)\right) 
        + \lambda \mathcal{R}\left(\mathbf{x}^j\right) \right. \\ &\left. 
        - \beta \sum_{i=1}^{S^2} \log \left\vert \frac{\partial \mathbf{x}_i^j}{\partial \mathbf{z}_{0}^j} \right\vert \right\},
    \end{aligned}
    \label{eq:optimization}
\end{equation}
where $\mathbf{A}(\cdot)$ is the forward model, $\mathbf{x}^j = \mathcal{PS} \left( \left\{ \mathrm{CNF}_\theta(\mathbf{z}^j_0,\mathbf{h}_i) \right\}_{i=1,\dots, S^2} \right)$, $\mathbf{z}_{0}^j \sim \mathcal{N}\left(0,\mathbf{I}_{N^2/S^2}\right)$, $M$ is the number of Monte-Carlo samples, and $\lambda, \beta$ are the weight parameters controlling the strength of the prior and entropy term. 
To demonstrate the flexibility of our method, we use a surrogate score prior derived by~\cite{feng2024variational} (the score model weights are frozen during optimization) and a classic TV prior as our explicit prior term.
For the score prior, we set $\lambda=1$ and $\beta=1$ in our experiment, given the fact that it's derived from the exact log probability likelihood.
For the TV prior, we empirically tuned $\lambda$ for each method.
However, the calibration of $\beta$ for an accurate uncertainty and the strength of prior $\lambda$ for a high image quality remains an open problem, which we discuss in Sec.~\ref{supp:ablations}. 
See red arrows in Fig.~\ref{fig:teaser}a for the path of gradient backpropagation from our three-part loss functions through the ShuffleFlow framework.
\begin{figure}[b]
    \centering
    \includegraphics[width=\linewidth]{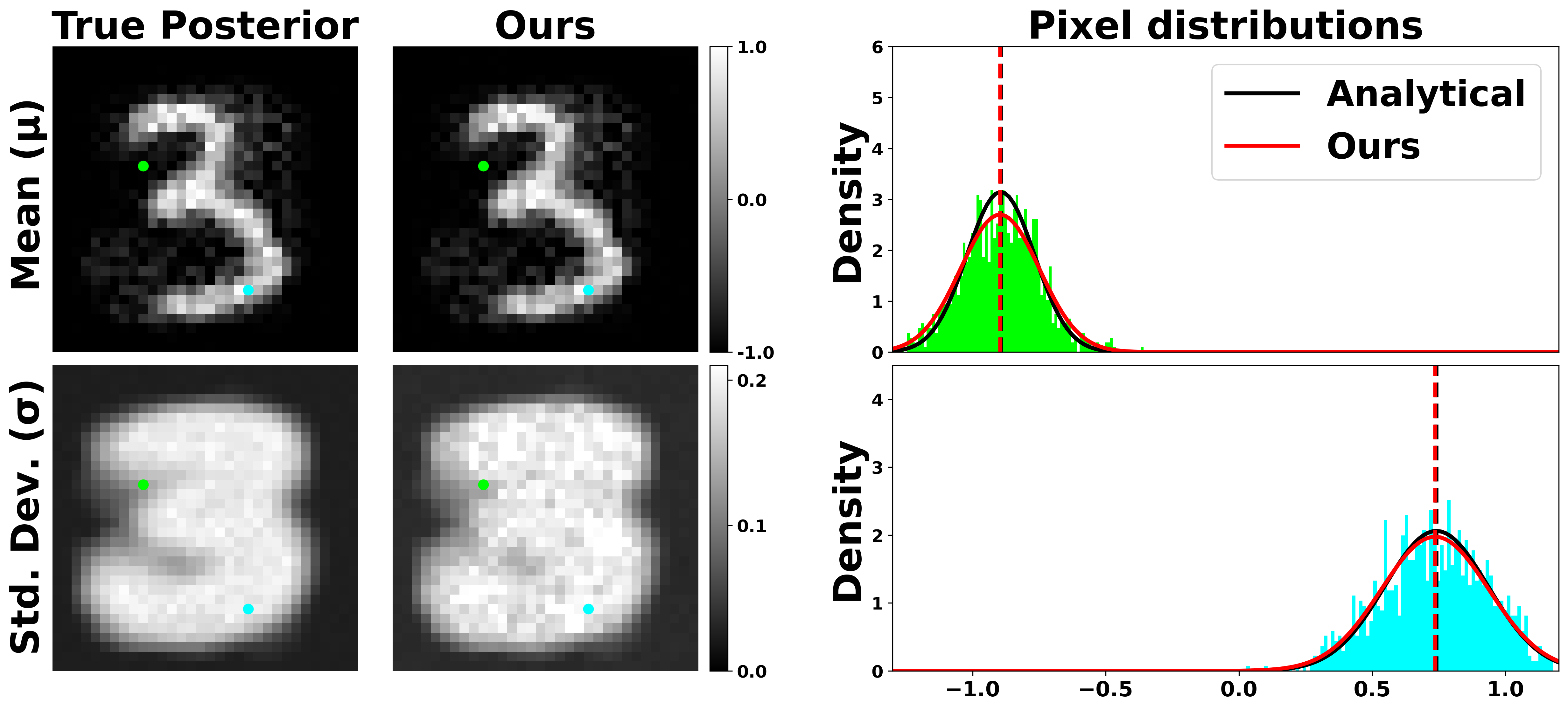}
    \caption{{\bf Gaussian toy example with analytical posterior.}
    We show that the posterior learned by ShuffleFlow matches the analytical posterior on the level of mean and standard deviation (left) as well as pixel value distributions (right). Vertical dashed lines show the mean pixel values.
    }
    \label{fig:gaussian_toy}
\end{figure}

\subsection{Inference}

After training, we calculate embeddings $\mathbf{h}_i$ using the NF encoder, sample several base variables $\mathbf{z}_0 \sim \mathcal{N}\left(0,\mathbf{I}_{N^2/S^2}\right)$, and feed them into ShuffleFlow to obtain posterior samples. 
Black arrows in Fig.~\ref{fig:teaser}a show forward sampling.

\begin{table*}[t] 
    \footnotesize
    \setlength{\tabcolsep}{3pt} 
    \centering
    \caption{
        {\bf Quantitative results for linear inverse problems with a pretrained score prior.}
        We compare to VI methods: DPI, CF-NeRF, and ours w/ patch, among which we are simultaneously SOTA in image quality and NLL. We also include diffusion sampling methods: Score-ALD, DPS, RED-Diff, PnP-DM, and DAPS. Note that the best diffusion samplers outperform our method, but scale poorly (see Fig.~\ref{fig:scale}a).
    }
    \label{tab:linear}
    
    \begin{tabular}{@{}lcccccccccccc@{}}
        \toprule
        \multirow{2}{*}{\bf Method} 
        & \multicolumn{6}{c}{\bf Motion Deblurring} 
        & \multicolumn{6}{c}{\bf Compressed Sensing MRI} \\
        \cmidrule(lr){2-7} \cmidrule(lr){8-13} 
        & \bf PSNR $\mathbf{\uparrow}$ & \bf SSIM $\mathbf{\uparrow}$ & \bf LPIPS $\mathbf{\downarrow}$ & \bf NLL $\mathbf{\downarrow}$ & \bf ECE $\mathbf{\downarrow}$ & \bf OODR $\mathbf{\downarrow}$
        & \bf PSNR $\mathbf{\uparrow}$ & \bf SSIM $\mathbf{\uparrow}$ & \bf LPIPS $\mathbf{\downarrow}$ & \bf NLL $\mathbf{\downarrow}$ & \bf ECE $\mathbf{\downarrow}$ & \bf OODR $\mathbf{\downarrow}$ \\
        \midrule
        DPI~\cite{sun2021deep}
        & 25.82 & 0.583 & 0.419 & -1.50 & 0.083 & 0.80\%
        & 33.51 & 0.813 & 0.184 & -2.42 & 0.041 & 0.59\%
        \\
        CF-NeRF~\cite{shen2022conditional}
        & 27.78 & 0.766 & 0.319 & -1.48 & 0.076 & 4.19\%
        & 34.51 & 0.857 & 0.183 & -2.45 & 0.051 & 2.75\%
        \\
        Ours w/ patch
        & 26.45 & 0.677 & 0.401 & -1.54 & 0.092 & 1.55\%
        & 29.18 & 0.725 & 0.274 & -1.74 & 0.051 & 4.36\%
        \\
        Ours
        & 28.36 & 0.767 & 0.325 & -1.74 & 0.099 & 2.71\%
        & 34.56 & 0.861 & 0.173 & -2.52 & 0.044 & 2.44\%
        \\
        \midrule 
        Score-ALD~\cite{jalal2021robust}
        & 25.88 & 0.589 & 0.370 & -1.64 & 0.028 & 0.68\%
        & 35.62 & 0.876 & 0.162 & -2.83 & 0.031 & 0.48\%
        \\
        DPS~\cite{chung2023diffusion}
        & 28.41 & 0.774 & 0.283 & -2.12 & 0.043 & 2.17\%
        & 34.76 & 0.863 & 0.216 & -2.76 & 0.039 & 0.88\%
        \\
        RED-Diff~\cite{mardani2024a}
        & 26.99 & 0.731 & 0.352 & -0.36 & 0.130 & 16.28\%
        & 33.75 & 0.843 & 0.279 & -2.04 & 0.108 & 9.79\%
        \\
        PnP-DM~\cite{wu2024principled} 
        & 28.00 & 0.725 & 0.285 & -1.82 & 0.068 & 3.55\%
        & 35.87 & 0.883 & 0.188 & -2.86 & 0.027 & 1.11 \%
        \\
        DAPS~\cite{zhang2025improving} 
        & 29.24 & 0.793 & 0.277 & -1.86 & 0.077 & 6.70\%
        & 34.96 & 0.864 & 0.162 & -2.71 & 0.025 & 1.10\%
        \\
        \bottomrule
    \end{tabular}
\end{table*}

\begin{figure*}
    \centering
    \includegraphics[width=\linewidth]{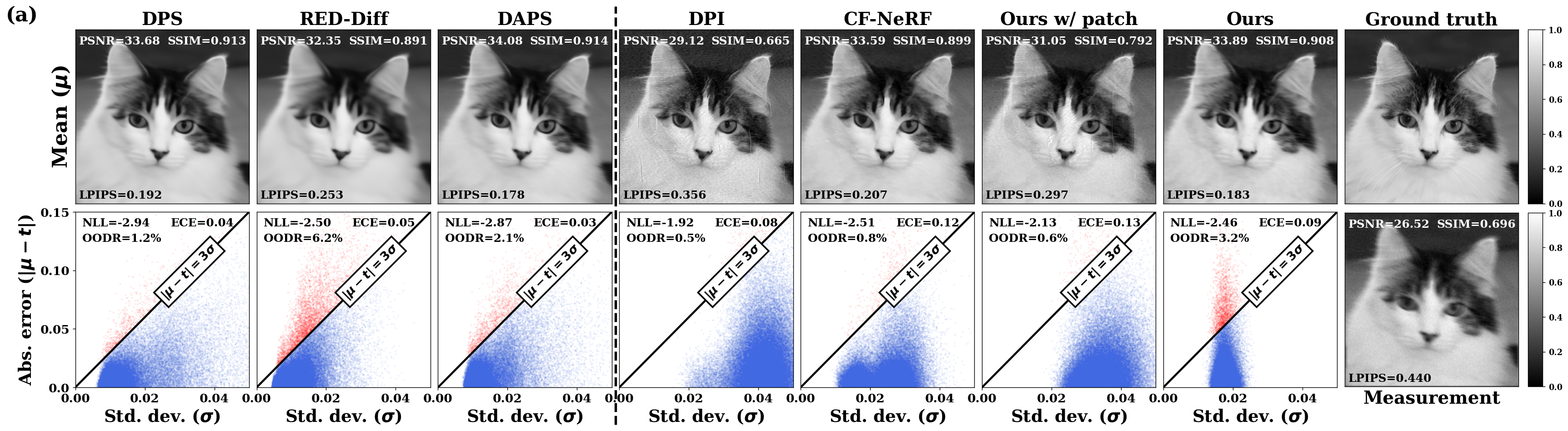}
    \includegraphics[width=\linewidth]{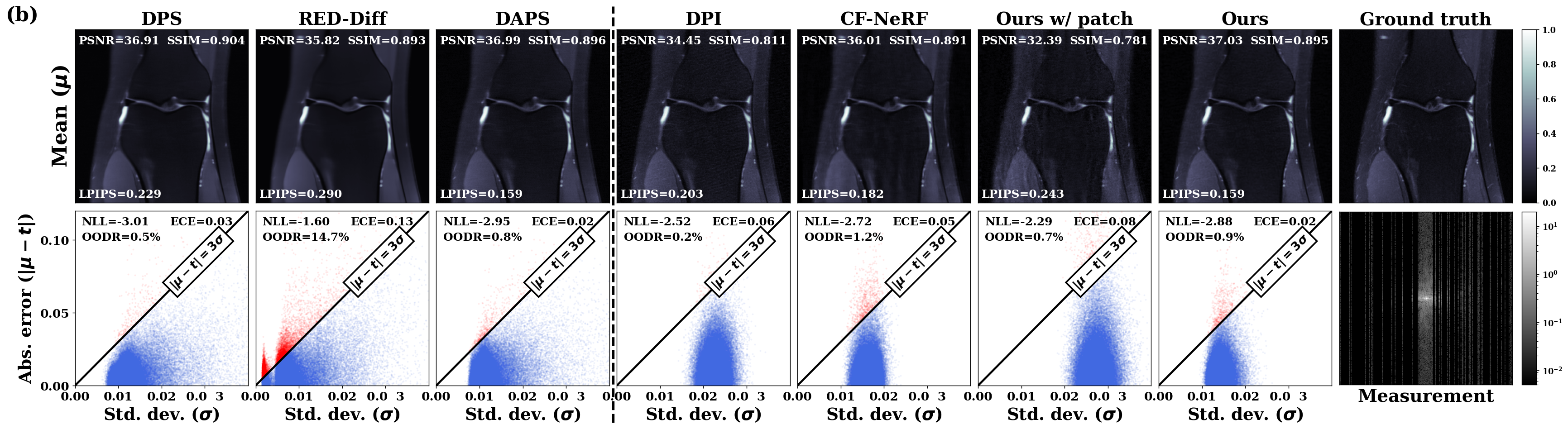}
    \caption{
        {\bf Visual samples on linear inverse problems with score prior: motion deblurring (a) and compressed sensing MRI (b).}
        We show the mean image computed from 128 posterior samples in row one and the scatter plot of absolute pixel errors vs. standard deviations in row two.
        Our method shows superior performance on image quality and uncertainty among VI methods and is comparable to the SOTA diffusion sampler DAPS~\cite{zhang2025improving}. We also illustrate results with TV prior in Fig.~\ref{fig:linear_tv}, and PnP-DM~\cite{wu2024principled} results in Fig.~\ref{fig:pnp-dm}.
        Error maps and individual samples are shown in Fig.~\ref{fig:linear_supp}.
    }
    \label{fig:linear_score}
\end{figure*}

\subsection{Implementation Details}

To implement our method, we use a conditional RealNVP~\cite{dinh2016density} as the backbone conditional normalizing flow network. 
We use SIREN~\cite{sitzmann2020implicit} as the NF encoder to calculate 128-dimensional embedding vectors $\mathbf{h}_i$.
In every Affine Coupling layer in the RealNVP, the embeddings are concatenated with the flow variable inputs and fed into a 2-layer MLP with LeakyReLU activation function to estimate the affine transform parameters~\cite{sun2021deep}.
We also applied an activation normalization layer~\cite{kingma2018glow} in place of a batch normalization layer before each flow function to improve the performance.
We used NCSN++~\cite{song2020score} as the score network and trained it with the variance preserving (VP) SDE with $\beta_\text{min}=0.1$ and $\beta_\text{max}=20$.
We set the Monte Carlo sample size $M=32$ for all our experiments.
See ablation studies on loss weights in Eq.~\ref{eq:optimization} and network architectures in Sec.~\ref{supp:ablations}.
All experiments were completed on two NVIDIA RTX A6000 GPUs each with 48 GB of VRAM.

\section{Experiments}
\label{sec:results}

\subsection{Metrics}
\label{sec:metrics}

To assess the accuracy of the reconstructed image and uncertainty, we select the following metrics: peak signal-to-noise ratio (PSNR $\uparrow$), structural similarity index measure (SSIM $\uparrow$), and learned perceptual image patch similarity (LPIPS $\downarrow$) for image quality, expected calibration error (ECE $\downarrow$) and negative log-likelihood (NLL $\downarrow$) for uncertainty~\cite{vasconcelos2022uncertainr}.
For visualizations in Fig.~\ref{fig:linear_score}, we show the portion of pixels falling outside of three standard deviations (absolute Z-score above 3) from ground truth~\cite{sun2021deep}, referred to as out-of-distribution rate (OODR $\downarrow$), which we also use as a metric for uncertainty calibration in this paper.
Note that these metrics only provide approximate evaluations of the uncertainty and both NLL and OODR should be interpreted with the context that true image posteriors do not follow pixel-wise independent Gaussian distributions. 

\begin{figure*}[t]
    \centering
    \includegraphics[width=0.97\columnwidth]{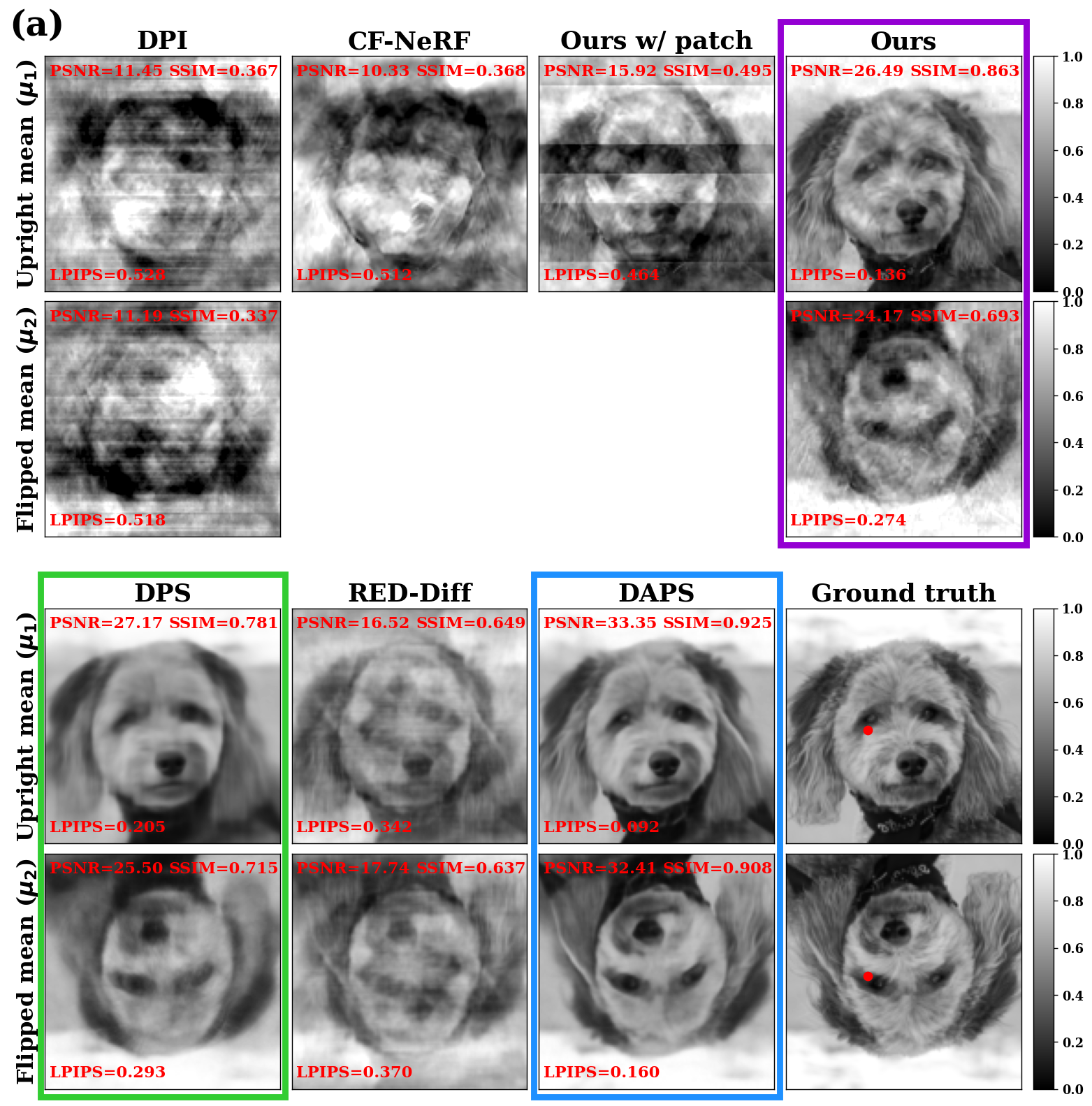}
    \hfill
    \includegraphics[width=1.02\columnwidth]{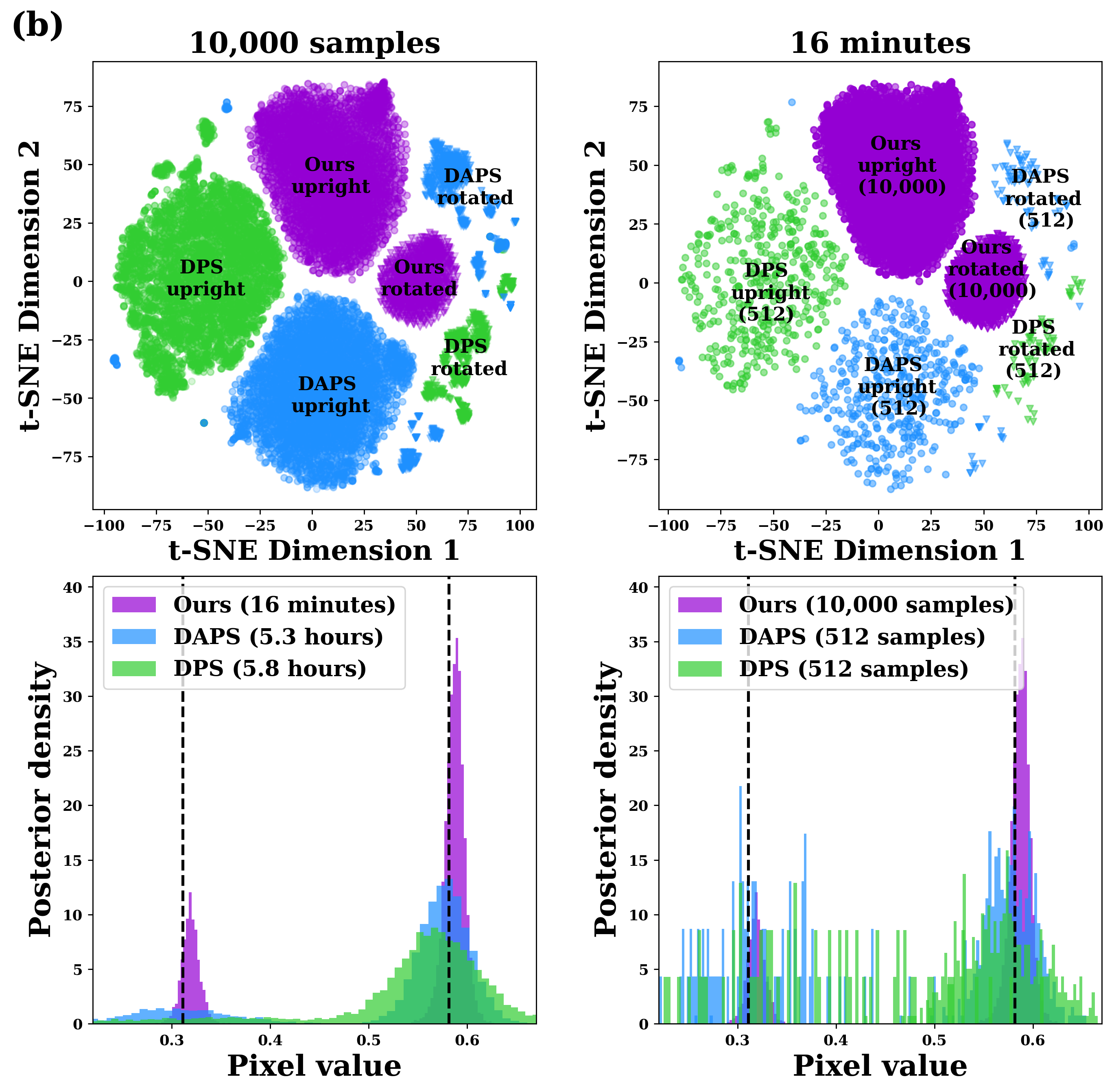}
    \caption{
        {\bf Results on nonlinear Fourier phase retrieval.}
        (a) We draw 128 samples from each method and filter the samples into different modes if a bimodal posterior exists.
        We show the mean image from each mode, and PSNR, SSIM, and LPIPS are calculated with respect to the ground truth of each mode.
        Our method correctly captures the bimodal posterior among VI methods with the best image quality.
        DAPS~\cite{zhang2025improving} stands out with the best image quality among all.
        (b) We compare our method with DAPS~\cite{zhang2025improving} and DPS~\cite{chung2023diffusion} in a large sample count (10,000, left) and a small inference time (16 min, right) scenario, showing the t-SNE visualization for the posterior samples (top) and pixel histograms (bottom).
        Our method is able to generate 10,000 posterior samples in 16 minutes (including both training and sampling time), resulting in a high-resolution bimodal posterior density in the histogram.
        In contrast, DAPS and DPS generate spiky and noisy posterior densities in a similar time and need 20$\times$ more time to generate a high-resolution posterior.
        See PnP-DM results in Fig.~\ref{fig:pnp-dm}.
    }
    \label{fig:fpr}
\end{figure*}

\subsection{Toy example}
\label{sec:toy_example}

We show a toy example with an analytically available Gaussian posterior, arising from a likelihood and prior that are both Gaussian. 
Specifically, we corrupt an image of digit ``3" from the MNIST~\cite{deng2012mnist} dataset with a compressed sensing matrix and additive Gaussian noise to create a Gaussian likelihood. 
We compute a Gaussian prior from all digit ``3" images in the MNIST dataset.
Fig.~\ref{fig:gaussian_toy} illustrates that we successfully capture the mean image as well as the standard deviations at each pixel. 
Note that the mean image has less noise in the periphery, where the prior has a lower standard deviation, reflecting the stronger regularization effect.
More details of this experiment are included in Sec.~\ref{supp:inverse_problems}.1.

\begin{table*}[t] 
    \footnotesize
    \setlength{\tabcolsep}{3pt} 
    \centering
    \caption{
        {\bf Results for nonlinear Fourier phase retrieval with a pretrained score prior.}
        We draw 128 samples for each method and classify them into two modes (PSNR $>5$ compared to upright or rotated ground truth) and compute the image quality metrics on the mean images for each mode. 
    We again outperform the other VI methods in image quality and NLL, but note that VI methods are systematically overconfident in this task, see Figs.~\ref{fig:fpr}b \& \ref{fig:fpr_supp}b. 
        Our problem decomposition allows our method to recover bimodality with fewer resources than DPI.
    }
    \label{tab:fpr}

        \begin{tabular}{@{}lcccccccccccccc@{}}
        \toprule
        \multirow{2}{*}{\bf Method} 
        & \multicolumn{7}{c}{\bf Upright mode} 
        & \multicolumn{7}{c}{\bf Rotated mode} \\
        \cmidrule(lr){2-8} \cmidrule(lr){9-15} 
        & \bf PSNR $\mathbf{\uparrow}$ & \bf SSIM $\mathbf{\uparrow}$ & \bf LPIPS $\mathbf{\downarrow}$ & \bf NLL $\mathbf{\downarrow}$ & \bf ECE $\mathbf{\downarrow}$ & \bf OODR $\mathbf{\downarrow}$ & \bf Pct.
        & \bf PSNR $\mathbf{\uparrow}$ & \bf SSIM $\mathbf{\uparrow}$ & \bf LPIPS $\mathbf{\downarrow}$ & \bf NLL $\mathbf{\downarrow}$ & \bf ECE $\mathbf{\downarrow}$ & \bf OODR $\mathbf{\downarrow}$ & \bf Pct. \\
        \midrule
        DPI~\cite{sun2021deep}
        & 15.55 & 0.366 & 0.493 & 202.52 & 0.384 & 79.4\% & 38.1\%
        & 15.10 & 0.360 & 0.488 & 222.82 & 0.384 & 81.8\% & 36.3\%
        \\
        CF-NeRF~\cite{shen2022conditional}
        & 12.28 & 0.279 & 0.510 & 663.54 & 0.274 & 90.6\% & 100.0\%
        & N/A & N/A & N/A & N/A & N/A & N/A & 0.0\%
        \\
        Ours w/ patch
        & 13.09 & 0.293 & 0.524 & 459.23 & 0.268 & 89.5\% & 100.0\%
        & N/A & N/A & N/A & N/A & N/A & N/A & 0.0\%
        \\
        Ours
        & 19.65 & 0.583 & 0.351 & 107.38 & 0.444 & 75.5\% & 53.6\%
        & 16.39 & 0.441 & 0.425 & 174.79 & 0.434 & 82.0\% & 38.4\%
        \\
        \midrule 
        Score-ALD~\cite{jalal2021robust,zheng2025inversebench}
        & 25.71 & 0.902 & 0.095 & -1.36 & 0.208 & 0.0\% & 57.5\%
        & 22.00 & 0.751 & 0.243 & -1.07 & 0.324 & 0.0\% & 42.4\%
        \\
        DPS~\cite{chung2023diffusion}
        & 21.93 & 0.624 & 0.326 & -1.16 & 0.390 & 0.5\% & 65.4\%
        & 18.48 & 0.508 & 0.428 & -0.72 & 0.378 & 1.4\% & 34.5\%
        \\
        RED-Diff~\cite{mardani2024a}
        & 17.84 & 0.589 & 0.380 & -0.64 & 0.423 & 0.4\% & 48.9\%
        & 17.94 & 0.579 & 0.390 & -0.65 & 0.423 & 0.4\% & 49.5\%
        \\
        PnP-DM~\cite{wu2024principled} 
        & 29.62 & 0.919 & 0.082 & -1.75 & 0.241 & 0.4\% & 70.5\%
        & 23.56 & 0.762 & 0.226 & -1.31 & 0.327 & 0.2\% & 29.3\%
        \\
        DAPS~\cite{zhang2025improving} 
        & 29.84 & 0.898 & 0.106 & -1.88 & 0.303 & 0.0\% & 79.5\%
        & 24.42 & 0.741 & 0.251 & 239.16 & 0.326 & 2.4\% & 20.5\%
        \\
        \bottomrule
    \end{tabular}
\end{table*}

\subsection{Linear inverse problems}
\label{sec:linear}

We test our method on two linear inverse problems: (1) motion deblurring with a kernel size of 61$\times$61 and intensity of 0.5, (2) compressed sensing MRI image reconstruction with Cartesian sampling masks and an acceleration rate of 4.0.
We use the AFHQ-v2 dataset~\cite{choi2020stargan} for motion deblurring and the fastMRI knee dataset~\cite{zbontar2018fastmri} for MRI. 
All images are resized to 256$\times$256 and normalized to [0,1].
See more details of the inverse problem setup in Sec.~\ref{supp:inverse_problems}.
We compare our method with several diffusion sampling methods: Score-ALD~\cite{jalal2021robust}, DPS~\cite{chung2023diffusion}, RED-Diff~\cite{mardani2024a}, PnP-DM~\cite{wu2024principled}, DAPS~\cite{zhang2025improving} (current SOTA according to~\cite{zheng2025inversebench}), and several VI methods: DPI~\cite{sun2021deep,feng2024variational}, CF-NeRF~\cite{shen2022conditional}.
As an ablation study, we also compare our method with a naive cropping strategy as in~\cite{meng2020gaussianization} (but share the CNF via conditioning like ours), which we refer to as ``Ours w/ patch" in this paper.
To use CF-NeRF for inverse imaging in our experiments, we reduce its flow dimension from 3 (for RGB images) to 1 (for grayscale images), and train with our three-component KL-divergence loss instead of the 3D reconstruction loss.
Note that in practice, we cannot set $S=256$ in ShuffleFlow to exactly recover CF-NeRF (1-dimensional), because the Affine Coupling layers require a minimum dimensionality of 2 due to their split-in-half operations.
To avoid such conflict, we use a 1-dimensional planar flow function in CF-NeRF.
For fair comparisons, all methods share the same pretrained score model. We scale down the layer size of the affine-coupling layers of DPI to fit it in our GPU memory.
With each method, we draw 128 posterior samples and calculate the mean image and pixel-wise standard deviations.
We report PSNR, SSIM, and LPIPS on the mean image and compute NLL, ECE, and OODR using the pixel-wise uncertainty.

We provide a quantitative analysis averaged over our test dataset in Tab.~\ref{tab:linear} and provide comparisons of visual samples of the mean image along with the scatter plot of absolute error vs. uncertainty for all sample image pixels in Fig.~\ref{fig:linear_score}.
CF-NeRF, at one extreme of VI sampling, can only model a posterior constrained to a one-dimensional manifold.
Though this only leads to slightly decreased image quality in linear problems since the posterior is single-mode, it is fundamentally unable to represent a bimodal posterior (see Sec.~\ref{sec:fpr}).
The shortcomings of local sampling are more apparent when spread across a patch in ours w/ patch, which displays extended spatial artifacts (see vertical stripes in Fig.~\ref{fig:linear_score}a).
On the other extreme of VI sampling, DPI exhaustively models the full posterior and generates reasonable image quality and uncertainty, at the cost of significantly increased time and memory.
In contrast, ShuffleFlow efficiently decomposes the problem and maintains global correlation, resulting in good image reconstruction and uncertainty estimation. We also include quantitative and qualitative results of motion deblurring and compressed sensing MRI with classic TV prior in Sec.~\ref{supp:results}, showing that our method also outperforms VI methods in the data-scarce setting.

As for diffusion samplers, Score-ALD generates high-quality images on MRI but suffers from noisy reconstruction in motion deblurring.
RED-Diff reconstructs slightly blurry images, which is due to its limited posterior expressiveness~\cite{mardani2024a}.
DPS, DAPS, and PnP-DM (see Fig.~\ref{fig:pnp-dm}) reconstruct good image quality on both tasks, while the latter has superior performance among diffusion samples.
This aligns with the observations in Inversebench~\cite{zheng2025inversebench}.

\subsection{Fourier phase retrieval}
\label{sec:fpr}

We further test our method on a {\em nonlinear} Fourier phase retrieval problem using 128$\times$128 images in the AFHQ-v2 dataset~\cite{choi2020stargan}.
Since this is a very challenging and ill-posed problem, we apply oversampling by zero-padding the image to 2$\times$ of its original size before calculating the Fourier space amplitude and add i.i.d. Gaussian noise with $\sigma=0.01$ to the measurements. 
See the detailed problem setup in Sec.~\ref{supp:inverse_problems}.4.
Note that due to the loss of Fourier phase information, an 180$^\circ$-rotated image would fit the measurement equally well as the upright image, leading to a bimodal posterior. 
We compare our method with DPI~\cite{sun2021deep}, CF-NeRF~\cite{shen2022conditional}, and ours w/ patch, and add 8 additional flow layers for each method to increase expressiveness.
We use Score-ALD~\cite{jalal2021robust}, DPS~\cite{chung2023diffusion}, RED-Diff~\cite{mardani2024a}, PnP-DM~\cite{wu2024principled}, DAPS~\cite{zhang2025improving} as baselines for comparison to diffusion samplers.
We draw 128 samples for each method and classify the samples into two modes based on whether they are more similar to the upright or rotated ground truth image (those with PSNR$<5$ in both situations are considered ``corrupted"). We then calculate the mean images separately for each mode. 

We provide quantitative analysis in Tab.~\ref{tab:fpr} and show visual samples of mean images in Fig.~\ref{fig:fpr}a. 
Among the VI methods, ours and DPI are able to capture a bimodal posterior, while CF-NeRF and ours w/ patch fail due to their inherent inability to model complex global correlations. 
Among the diffusion samplers, DAPS and PnP-DM (see visual samples in Fig.~\ref{fig:pnp-dm}c) capture a very high-resolution posterior for both modes, while DPS reconstructions show slight blurring.
RED-Diff has a very corrupted bimodal posterior, likely due to its limited posterior expressiveness~\cite{mardani2024a}.
To our knowledge, ours is the first VI-based method that captures a bimodal posterior with the surrogate score prior~\cite{feng2024variational} in an inverse imaging problem. Tab.~\ref{tab:fpr} shows dramatically reduced UQ performance in VI methods compared to diffusion sampling methods. We note that this is due to systematic overconfidence, see Figs.~\ref{fig:fpr}b \& \ref{fig:fpr_supp}b. Overconfidence can be addressed by increasing the entropy weight, but we observe that this currently leads to mode collapse (see Fig.~\ref{fig:fpr_beta}), and must be addressed by future work that stabilizes training.

In Fig.~\ref{fig:fpr}b, we compare DAPS, DPS, and ours, as representative methods, 
under two settings: 1. a \textit{high-resolution} posterior with 10,000 posterior samples (left), and a 16-minute short time budget (right), which are equivalent for our method but not for the diffusion samplers.
We provide the t-SNE visualization of the posterior samples, select a pixel around the dog's eye with clear bimodal values (marked with red dots in Fig.~\ref{fig:fpr}a), and show the pixel value histogram from the posterior samples.
The vertical dashed lines show the ground truth pixel values for each mode.
Our posterior samples reveal a clear bimodal posterior in both the t-SNE visualizations and the pixel histograms.
In the time-limited setting, DAPS and DPS can only generate 512 posterior samples, leading to a noisy pixel histogram, and are unable to clearly characterize the second mode.
This mode begins to be recovered by all three methods in the 10,000-sample setting, but DAPS and DPS need 20$\times$ more sampling time and show less clear structure in t-SNE projection.
This demonstrates a significant advantage of our method over diffusion samplers when a high-resolution posterior is needed in scientific applications --- the former can generate large numbers of samples cheaply after a short training time, while the latter's computational costs grow linearly with the number of samples needed.

\section{Conclusion}
\label{sec:conclusion}

We have introduced ShuffleFlow, an efficient NF+CNF framework for scalable posterior estimation in Bayesian inverse imaging. 
We break down the problem of modeling a full image posterior into three parts: coordinate sampling (pixel-unshuffling), feature encoding (NF), and posterior estimation (CNF), ShuffleFlow efficiently models the posterior as the joint distribution of the downsampled image stack. We demonstrate theoretically and empirically that this decomposition significantly reduces the computational resources while maintaining posterior expressiveness.

Our experiments show that ShuffleFlow, among VI methods, is simultaneously SOTA on image quality, uncertainty calibration, and computational scalability. 
We further show our method's ability to quickly discover bimodality in the posterior of the nonlinear Fourier phase retrieval problem, 20$\times$ faster than the best diffusion samplers. This suggests that VI methods like ours can serve a vital role alongside diffusion samplers, as exploratory tools to determine when posterior complexity justifies extensive sample generation (Fig.~\ref{fig:fpr}b) and in low data regimes where diffusion models cannot be trained (Sec.~\ref{supp:results}.1).

A major current shortcoming is overconfidence resulting from training instability in our nonlinear problem example (see Sec.~\ref{supp:results}.3). 
Future work should increase the performance with attention-based coordinate sampling strategy, normalizing flow architecture, and generative priors (e.g.,~\cite{erbach2025solving}).
Another future direction is applying our method to scientific applications (e.g.,~\cite{karchev2022strong,adam2022posterior,feng2024event,barco2025blind}), where a high-resolution posterior is required for downstream science.

\ifpeerreview \else
\section*{Acknowledgments}
We gratefully acknowledge the support of the NSF-Simons AI Institute for the Sky (SkAI) via grants NSF AST-2421845 and Simons Foundation MPS-AI-00010513.
This material is based upon work supported by the U.S. National Science Foundation under Award No. 2542022.
The authors would like to thank Bryan Pardo, He Sun, and Yi-Chun Hung for helpful discussions. 
\fi

\bibliographystyle{IEEEtran}
\bibliography{references}

@String(ICLR = {Int. Conf. Learn. Represent.})

@String(AAAI = {AAAI})

@String(ICLR  = {ICLR})

@inproceedings{shi2016real,
  title={Real-time single image and video super-resolution using an efficient sub-pixel convolutional neural network},
  author={Shi, Wenzhe and Caballero, Jose and Husz{\'a}r, Ferenc and Totz, Johannes and Aitken, Andrew P and Bishop, Rob and Rueckert, Daniel and Wang, Zehan},
  booktitle={Proceedings of the IEEE conference on computer vision and pattern recognition},
  pages={1874--1883},
  year={2016}
}

@article{sun2021scalable,
  title={Scalable plug-and-play ADMM with convergence guarantees},
  author={Sun, Yu and Wu, Zihui and Xu, Xiaojian and Wohlberg, Brendt and Kamilov, Ulugbek S},
  journal={IEEE Transactions on Computational Imaging},
  volume={7},
  pages={849--863},
  year={2021},
  publisher={IEEE}
}

@inproceedings{sajjadi2018frame,
  title={Frame-recurrent video super-resolution},
  author={Sajjadi, Mehdi SM and Vemulapalli, Raviteja and Brown, Matthew},
  booktitle={Proceedings of the IEEE conference on computer vision and pattern recognition},
  pages={6626--6634},
  year={2018}
}

@inproceedings{gu2019self,
  title={Self-guided network for fast image denoising},
  author={Gu, Shuhang and Li, Yawei and Gool, Luc Van and Timofte, Radu},
  booktitle={Proceedings of the IEEE/CVF International Conference on Computer Vision},
  pages={2511--2520},
  year={2019}
}

@inproceedings{huang2021neighbor2neighbor,
  title={Neighbor2neighbor: Self-supervised denoising from single noisy images},
  author={Huang, Tao and Li, Songjiang and Jia, Xu and Lu, Huchuan and Liu, Jianzhuang},
  booktitle={Proceedings of the IEEE/CVF conference on computer vision and pattern recognition},
  pages={14781--14790},
  year={2021}
}

@inproceedings{mansour2023zero,
  title={Zero-shot noise2noise: Efficient image denoising without any data},
  author={Mansour, Youssef and Heckel, Reinhard},
  booktitle={Proceedings of the IEEE/CVF Conference on Computer Vision and Pattern Recognition},
  pages={14018--14027},
  year={2023}
}

@article{cifarelli1996finetti,
  title={De Finetti's contribution to probability and statistics},
  author={Cifarelli, Donato Michele and Regazzini, Eugenio},
  journal={Statistical Science},
  pages={253--282},
  year={1996},
  publisher={JSTOR}
}

@article{sitzmann2020implicit,
  title={Implicit neural representations with periodic activation functions},
  author={Sitzmann, Vincent and Martel, Julien and Bergman, Alexander and Lindell, David and Wetzstein, Gordon},
  journal={Advances in neural information processing systems},
  volume={33},
  pages={7462--7473},
  year={2020}
}

@inproceedings{xie2022neural,
  title={Neural fields in visual computing and beyond},
  author={Xie, Yiheng and Takikawa, Towaki and Saito, Shunsuke and Litany, Or and Yan, Shiqin and Khan, Numair and Tombari, Federico and Tompkin, James and Sitzmann, Vincent and Sridhar, Srinath},
  booktitle={Computer Graphics Forum},
  volume={41},
  number={2},
  pages={641--676},
  year={2022},
  organization={Wiley Online Library}
}

@inproceedings{kimgrids,
  title={Grids Often Outperform Implicit Neural Representation at Compressing Dense Signals},
  author={Kim, Namhoon and Fridovich-Keil, Sara},
  booktitle={The Thirty-ninth Annual Conference on Neural Information Processing Systems}
}

@article{essakinewe,
  title={Where Do We Stand with Implicit Neural Representations? A Technical and Performance Survey},
  author={Essakine, Amer and Cheng, Yanqi and Cheng, Chun-Wun and Zhang, Lipei and Deng, Zhongying and Zhu, Lei and Sch{\"o}nlieb, Carola-Bibiane and Aviles-Rivero, Angelica I},
  journal={Transactions on Machine Learning Research}
}

@article{mildenhall2021nerf,
  title={Nerf: Representing scenes as neural radiance fields for view synthesis},
  author={Mildenhall, Ben and Srinivasan, Pratul P and Tancik, Matthew and Barron, Jonathan T and Ramamoorthi, Ravi and Ng, Ren},
  journal={Communications of the ACM},
  volume={65},
  number={1},
  pages={99--106},
  year={2021},
  publisher={ACM New York, NY, USA}
}

@inproceedings{martin2021nerf,
  title={Nerf in the wild: Neural radiance fields for unconstrained photo collections},
  author={Martin-Brualla, Ricardo and Radwan, Noha and Sajjadi, Mehdi SM and Barron, Jonathan T and Dosovitskiy, Alexey and Duckworth, Daniel},
  booktitle={Proceedings of the IEEE/CVF conference on computer vision and pattern recognition},
  pages={7210--7219},
  year={2021}
}

@inproceedings{barron2021mip,
  title={Mip-nerf: A multiscale representation for anti-aliasing neural radiance fields},
  author={Barron, Jonathan T and Mildenhall, Ben and Tancik, Matthew and Hedman, Peter and Martin-Brualla, Ricardo and Srinivasan, Pratul P},
  booktitle={Proceedings of the IEEE/CVF international conference on computer vision},
  pages={5855--5864},
  year={2021}
}

@inproceedings{molaei2023implicit,
  title={Implicit neural representation in medical imaging: A comparative survey},
  author={Molaei, Amirali and Aminimehr, Amirhossein and Tavakoli, Armin and Kazerouni, Amirhossein and Azad, Bobby and Azad, Reza and Merhof, Dorit},
  booktitle={Proceedings of the IEEE/CVF International Conference on Computer Vision},
  pages={2381--2391},
  year={2023}
}

@article{xu2023nesvor,
  title={NeSVoR: implicit neural representation for slice-to-volume reconstruction in MRI},
  author={Xu, Junshen and Moyer, Daniel and Gagoski, Borjan and Iglesias, Juan Eugenio and Grant, P Ellen and Golland, Polina and Adalsteinsson, Elfar},
  journal={IEEE transactions on medical imaging},
  volume={42},
  number={6},
  pages={1707--1719},
  year={2023},
  publisher={IEEE}
}

@article{shen2022nerp,
  title={NeRP: implicit neural representation learning with prior embedding for sparsely sampled image reconstruction},
  author={Shen, Liyue and Pauly, John and Xing, Lei},
  journal={IEEE Transactions on Neural Networks and Learning Systems},
  volume={35},
  number={1},
  pages={770--782},
  year={2022},
  publisher={IEEE}
}

@inproceedings{zang2021intratomo,
  title={Intratomo: self-supervised learning-based tomography via sinogram synthesis and prediction},
  author={Zang, Guangming and Idoughi, Ramzi and Li, Rui and Wonka, Peter and Heidrich, Wolfgang},
  booktitle={Proceedings of the IEEE/CVF International Conference on Computer Vision},
  pages={1960--1970},
  year={2021}
}

@article{sun2021coil,
  title={Coil: Coordinate-based internal learning for tomographic imaging},
  author={Sun, Yu and Liu, Jiaming and Xie, Mingyang and Wohlberg, Brendt and Kamilov, Ulugbek S},
  journal={IEEE Transactions on Computational Imaging},
  volume={7},
  pages={1400--1412},
  year={2021},
  publisher={IEEE}
}

@article{liu2022recovery,
  title={Recovery of continuous 3d refractive index maps from discrete intensity-only measurements using neural fields},
  author={Liu, Renhao and Sun, Yu and Zhu, Jiabei and Tian, Lei and Kamilov, Ulugbek S},
  journal={Nature Machine Intelligence},
  volume={4},
  number={9},
  pages={781--791},
  year={2022},
  publisher={Nature Publishing Group UK London}
}

@article{zhou2023fourier,
  title={Fourier ptychographic microscopy image stack reconstruction using implicit neural representations},
  author={Zhou, Haowen and Feng, Brandon Y and Guo, Haiyun and Lin, Siyu Steven and Liang, Mingshu and Metzler, Christopher A and Yang, Changhuei},
  journal={Optica},
  volume={10},
  number={12},
  pages={1679--1687},
  year={2023},
  publisher={Optica Publishing Group}
}

@article{cao2024neural,
  title={Neural space--time model for dynamic multi-shot imaging},
  author={Cao, Ruiming and Divekar, Nikita S and Nu{\~n}ez, James K and Upadhyayula, Srigokul and Waller, Laura},
  journal={Nature Methods},
  pages={1--6},
  year={2024},
  publisher={Nature Publishing Group US New York}
}

@inproceedings{li2025coordinate,
  title={Coordinate-based Speed of Sound Recovery for Aberration-Corrected Photoacoustic Computed Tomography},
  author={Li, Tianao and Cui, Manxiu and Ma, Cheng and Alexander, Emma},
  booktitle={Proceedings of the IEEE/CVF International Conference on Computer Vision},
  pages={27466--27475},
  year={2025}
}

@inproceedings{venkatakrishnan2013plug,
  title={Plug-and-play priors for model based reconstruction},
  author={Venkatakrishnan, Singanallur V and Bouman, Charles A and Wohlberg, Brendt},
  booktitle={2013 IEEE global conference on signal and information processing},
  pages={945--948},
  year={2013},
  organization={IEEE}
}

@article{romano2017little,
  title={The little engine that could: Regularization by denoising (RED)},
  author={Romano, Yaniv and Elad, Michael and Milanfar, Peyman},
  journal={SIAM Journal on Imaging Sciences},
  volume={10},
  number={4},
  pages={1804--1844},
  year={2017},
  publisher={SIAM}
}

@article{candes2007sparsity,
  title={Sparsity and incoherence in compressive sampling},
  author={Candes, Emmanuel and Romberg, Justin},
  journal={Inverse problems},
  volume={23},
  number={3},
  pages={969},
  year={2007},
  publisher={IOP Publishing}
}

@article{chan2016plug,
  title={Plug-and-play ADMM for image restoration: Fixed-point convergence and applications},
  author={Chan, Stanley H and Wang, Xiran and Elgendy, Omar A},
  journal={IEEE Transactions on Computational Imaging},
  volume={3},
  number={1},
  pages={84--98},
  year={2016},
  publisher={IEEE}
}

@article{monakhova2019learned,
  title={Learned reconstructions for practical mask-based lensless imaging},
  author={Monakhova, Kristina and Yurtsever, Joshua and Kuo, Grace and Antipa, Nick and Yanny, Kyrollos and Waller, Laura},
  journal={Optics express},
  volume={27},
  number={20},
  pages={28075--28090},
  year={2019},
  publisher={Optical Society of America}
}

@article{ho2020denoising,
  title={Denoising diffusion probabilistic models},
  author={Ho, Jonathan and Jain, Ajay and Abbeel, Pieter},
  journal={Advances in neural information processing systems},
  volume={33},
  pages={6840--6851},
  year={2020}
}

@article{song2020denoising,
  title={Denoising diffusion implicit models},
  author={Song, Jiaming and Meng, Chenlin and Ermon, Stefano},
  journal={arXiv preprint arXiv:2010.02502},
  year={2020}
}

@article{song2019generative,
  title={Generative modeling by estimating gradients of the data distribution},
  author={Song, Yang and Ermon, Stefano},
  journal={Advances in neural information processing systems},
  volume={32},
  year={2019}
}

@article{song2020score,
  title={Score-based generative modeling through stochastic differential equations},
  author={Song, Yang and Sohl-Dickstein, Jascha and Kingma, Diederik P and Kumar, Abhishek and Ermon, Stefano and Poole, Ben},
  journal={arXiv preprint arXiv:2011.13456},
  year={2020}
}

@inproceedings{
    mardani2024a,
    title={A Variational Perspective on Solving Inverse Problems with Diffusion Models},
    author={Morteza Mardani and Jiaming Song and Jan Kautz and Arash Vahdat},
    booktitle={The Twelfth International Conference on Learning Representations},
    year={2024},
    url={https://openreview.net/forum?id=1YO4EE3SPB}
}

@inproceedings{chung2023diffusion,
  title={DIFFUSION POSTERIOR SAMPLING FOR GENERAL NOISY INVERSE PROBLEMS},
  author={Chung, Hyungjin and Kim, Jeongsol and McCann, Michael T and Klasky, Marc L and Ye, Jong Chul},
  booktitle={11th International Conference on Learning Representations, ICLR 2023},
  year={2023}
}

@article{jalal2021robust,
  title={Robust compressed sensing mri with deep generative priors},
  author={Jalal, Ajil and Arvinte, Marius and Daras, Giannis and Price, Eric and Dimakis, Alexandros G and Tamir, Jon},
  journal={Advances in Neural Information Processing Systems},
  volume={34},
  pages={14938--14954},
  year={2021}
}

@article{graikos2022diffusion,
  title={Diffusion models as plug-and-play priors},
  author={Graikos, Alexandros and Malkin, Nikolay and Jojic, Nebojsa and Samaras, Dimitris},
  journal={Advances in Neural Information Processing Systems},
  volume={35},
  pages={14715--14728},
  year={2022}
}

@article{kawar2022denoising,
  title={Denoising diffusion restoration models},
  author={Kawar, Bahjat and Elad, Michael and Ermon, Stefano and Song, Jiaming},
  journal={Advances in Neural Information Processing Systems},
  volume={35},
  pages={23593--23606},
  year={2022}
}

@inproceedings{
    song2022solving,
    title={Solving Inverse Problems in Medical Imaging with Score-Based Generative Models},
    author={Yang Song and Liyue Shen and Lei Xing and Stefano Ermon},
    booktitle={International Conference on Learning Representations},
    year={2022},
    url={https://openreview.net/forum?id=vaRCHVj0uGI}
}

@article{chung2022score,
  title={Score-based diffusion models for accelerated MRI},
  author={Chung, Hyungjin and Ye, Jong Chul},
  journal={Medical image analysis},
  volume={80},
  pages={102479},
  year={2022},
  publisher={Elsevier}
}

@article{chung2022improving,
  title={Improving diffusion models for inverse problems using manifold constraints},
  author={Chung, Hyungjin and Sim, Byeongsu and Ryu, Dohoon and Ye, Jong Chul},
  journal={Advances in Neural Information Processing Systems},
  volume={35},
  pages={25683--25696},
  year={2022}
}

@article{wu2024principled,
  title={Principled probabilistic imaging using diffusion models as plug-and-play priors},
  author={Wu, Zihui and Sun, Yu and Chen, Yifan and Zhang, Bingliang and Yue, Yisong and Bouman, Katherine},
  journal={Advances in Neural Information Processing Systems},
  volume={37},
  pages={118389--118427},
  year={2024}
}

@article{sun2024provable,
  title={Provable probabilistic imaging using score-based generative priors},
  author={Sun, Yu and Wu, Zihui and Chen, Yifan and Feng, Berthy T and Bouman, Katherine L},
  journal={IEEE Transactions on Computational Imaging},
  year={2024},
  publisher={IEEE}
}

@inproceedings{zhang2025improving,
  title={Improving diffusion inverse problem solving with decoupled noise annealing},
  author={Zhang, Bingliang and Chu, Wenda and Berner, Julius and Meng, Chenlin and Anandkumar, Anima and Song, Yang},
  booktitle={Proceedings of the Computer Vision and Pattern Recognition Conference},
  pages={20895--20905},
  year={2025}
}

@article{blei2017variational,
  title={Variational inference: A review for statisticians},
  author={Blei, David M and Kucukelbir, Alp and McAuliffe, Jon D},
  journal={Journal of the American statistical Association},
  volume={112},
  number={518},
  pages={859--877},
  year={2017},
  publisher={Taylor \& Francis}
}

@inproceedings{rezende2015variational,
  title={Variational inference with normalizing flows},
  author={Rezende, Danilo and Mohamed, Shakir},
  booktitle={International conference on machine learning},
  pages={1530--1538},
  year={2015},
  organization={PMLR}
}

@article{papamakarios2021normalizing,
  title={Normalizing flows for probabilistic modeling and inference},
  author={Papamakarios, George and Nalisnick, Eric and Rezende, Danilo Jimenez and Mohamed, Shakir and Lakshminarayanan, Balaji},
  journal={Journal of Machine Learning Research},
  volume={22},
  number={57},
  pages={1--64},
  year={2021}
}

@article{kingma2018glow,
  title={Glow: Generative flow with invertible 1x1 convolutions},
  author={Kingma, Durk P and Dhariwal, Prafulla},
  journal={Advances in neural information processing systems},
  volume={31},
  year={2018}
}

@article{dinh2016density,
  title={Density estimation using real nvp},
  author={Dinh, Laurent and Sohl-Dickstein, Jascha and Bengio, Samy},
  journal={arXiv preprint arXiv:1605.08803},
  year={2016}
}

@inproceedings{dai2021sliced,
  title={Sliced Iterative Normalizing Flows},
  author={Dai, Biwei and Seljak, Uros},
  booktitle={International Conference on Machine Learning},
  pages={2352--2364},
  year={2021},
  organization={PMLR}
}

@inproceedings{meng2020gaussianization,
  title={Gaussianization flows},
  author={Meng, Chenlin and Song, Yang and Song, Jiaming and Ermon, Stefano},
  booktitle={International Conference on Artificial Intelligence and Statistics},
  pages={4336--4345},
  year={2020},
  organization={PMLR}
}

@inproceedings{yang2019pointflow,
  title={Pointflow: 3d point cloud generation with continuous normalizing flows},
  author={Yang, Guandao and Huang, Xun and Hao, Zekun and Liu, Ming-Yu and Belongie, Serge and Hariharan, Bharath},
  booktitle={Proceedings of the IEEE/CVF international conference on computer vision},
  pages={4541--4550},
  year={2019}
}

@inproceedings{pumarola2020c,
  title={C-flow: Conditional generative flow models for images and 3d point clouds},
  author={Pumarola, Albert and Popov, Stefan and Moreno-Noguer, Francesc and Ferrari, Vittorio},
  booktitle={Proceedings of the IEEE/CVF Conference on Computer Vision and Pattern Recognition},
  pages={7949--7958},
  year={2020}
}

@inproceedings{hong2023robustness,
  title={On the robustness of normalizing flows for inverse problems in imaging},
  author={Hong, Seongmin and Park, Inbum and Chun, Se Young},
  booktitle={Proceedings of the IEEE/CVF International Conference on Computer Vision},
  pages={10745--10755},
  year={2023}
}

@article{bardsley2012mcmc,
  title={MCMC-based image reconstruction with uncertainty quantification},
  author={Bardsley, Johnathan M},
  journal={SIAM Journal on Scientific Computing},
  volume={34},
  number={3},
  pages={A1316--A1332},
  year={2012},
  publisher={SIAM}
}

@article{cardoso2023monte,
  title={Monte Carlo guided diffusion for Bayesian linear inverse problems},
  author={Cardoso, Gabriel and Idrissi, Yazid Janati El and Corff, Sylvain Le and Moulines, Eric},
  journal={arXiv preprint arXiv:2308.07983},
  year={2023}
}

@book{neal2012bayesian,
  title={Bayesian learning for neural networks},
  author={Neal, Radford M},
  volume={118},
  year={2012},
  publisher={Springer Science \& Business Media}
}

@inproceedings{gal2016dropout,
  title={Dropout as a bayesian approximation: Representing model uncertainty in deep learning},
  author={Gal, Yarin and Ghahramani, Zoubin},
  booktitle={international conference on machine learning},
  pages={1050--1059},
  year={2016},
  organization={PMLR}
}

@article{lakshminarayanan2017simple,
  title={Simple and scalable predictive uncertainty estimation using deep ensembles},
  author={Lakshminarayanan, Balaji and Pritzel, Alexander and Blundell, Charles},
  journal={Advances in neural information processing systems},
  volume={30},
  year={2017}
}

@article{kendall2017uncertainties,
  title={What uncertainties do we need in bayesian deep learning for computer vision?},
  author={Kendall, Alex and Gal, Yarin},
  journal={Advances in neural information processing systems},
  volume={30},
  year={2017}
}

@article{laumont2022bayesian,
  title={Bayesian imaging using plug \& play priors: when langevin meets tweedie},
  author={Laumont, R{\'e}mi and Bortoli, Valentin De and Almansa, Andr{\'e}s and Delon, Julie and Durmus, Alain and Pereyra, Marcelo},
  journal={SIAM Journal on Imaging Sciences},
  volume={15},
  number={2},
  pages={701--737},
  year={2022},
  publisher={SIAM}
}

@inproceedings{pereyra2024equivariant,
  title={Equivariant bootstrapping for uncertainty quantification in imaging inverse problems},
  author={Pereyra, Marcelo and Tachella, Juli{\'a}n},
  booktitle={International Conference on Artificial Intelligence and Statistics},
  pages={4141--4149},
  year={2024},
  organization={PMLR}
}

@inproceedings{angelopoulos2022image,
  title={Image-to-image regression with distribution-free uncertainty quantification and applications in imaging},
  author={Angelopoulos, Anastasios N and Kohli, Amit Pal and Bates, Stephen and Jordan, Michael and Malik, Jitendra and Alshaabi, Thayer and Upadhyayula, Srigokul and Romano, Yaniv},
  booktitle={International Conference on Machine Learning},
  pages={717--730},
  year={2022},
  organization={PMLR}
}

@article{ye2025learned,
  title={Learned, uncertainty-driven adaptive acquisition for photon-efficient scanning microscopy},
  author={Ye, Cassandra Tong and Han, Jiashu and Liu, Kunzan and Angelopoulos, Anastasios and Griffith, Linda and Monakhova, Kristina and You, Sixian},
  journal={Optics Express},
  volume={33},
  number={6},
  pages={12269--12287},
  year={2025},
  publisher={Optica Publishing Group}
}

@article{xue2019reliable,
  title={Reliable deep-learning-based phase imaging with uncertainty quantification},
  author={Xue, Yujia and Cheng, Shiyi and
          Li, Yunzhe and Tian, Lei},
  journal={Optica},
  volume={6},
  number={5},
  pages={618--629},
  year={2019},
  publisher={Optica Publishing Group}
}

@article{repetti2019scalable,
  title={Scalable Bayesian uncertainty quantification in imaging inverse problems via convex optimization},
  author={Repetti, Audrey and Pereyra, Marcelo and Wiaux, Yves},
  journal={SIAM Journal on Imaging Sciences},
  volume={12},
  number={1},
  pages={87--118},
  year={2019},
  publisher={SIAM}
}

@inproceedings{feng2023score,
  title={Score-based diffusion models as principled priors for inverse imaging},
  author={Feng, Berthy T and Smith, Jamie and Rubinstein, Michael and Chang, Huiwen and Bouman, Katherine L and Freeman, William T},
  booktitle={Proceedings of the IEEE/CVF International Conference on Computer Vision},
  pages={10520--10531},
  year={2023}
}

@article{feng2024variational,
    title={Variational Bayesian Imaging with an Efficient Surrogate Score-based Prior},
    author={Berthy Feng and Katherine Bouman},
    journal={Transactions on Machine Learning Research},
    issn={2835-8856},
    year={2024},
    url={https://openreview.net/forum?id=db2pFKVcm1},
    note={}
}

@article{feng2024event,
  title={Event-horizon-scale Imaging of M87* under Different Assumptions via Deep Generative Image Priors},
  author={Feng, Berthy T and Bouman, Katherine L and Freeman, William T},
  journal={arXiv preprint arXiv:2406.02785},
  year={2024}
}

@inproceedings{sun2021deep,
  title={Deep probabilistic imaging: Uncertainty quantification and multi-modal solution characterization for computational imaging},
  author={Sun, He and Bouman, Katherine L},
  booktitle={Proceedings of the AAAI Conference on Artificial Intelligence},
  volume={35},
  number={3},
  pages={2628--2637},
  year={2021}
}

@article{sun2022alpha,
  title={$\alpha$-deep probabilistic inference ($\alpha$-dpi): efficient uncertainty quantification from exoplanet astrometry to black hole feature extraction},
  author={Sun, He and Bouman, Katherine L and Tiede, Paul and Wang, Jason J and Blunt, Sarah and Mawet, Dimitri},
  journal={The Astrophysical Journal},
  volume={932},
  number={2},
  pages={99},
  year={2022},
  publisher={IOP Publishing}
}

@article{vasconcelos2022uncertainr,
  title={Uncertainr: Uncertainty quantification of end-to-end implicit neural representations for computed tomography},
  author={Vasconcelos, Francisca and He, Bobby and Singh, Nalini and Teh, Yee Whye},
  journal={arXiv preprint arXiv:2202.10847},
  year={2022}
}

@inproceedings{shen2021stochastic,
  title={Stochastic neural radiance fields: Quantifying uncertainty in implicit 3d representations},
  author={Shen, Jianxiong and Ruiz, Adria and Agudo, Antonio and Moreno-Noguer, Francesc},
  booktitle={2021 International Conference on 3D Vision (3DV)},
  pages={972--981},
  year={2021},
  organization={IEEE}
}

@inproceedings{shen2022conditional,
  title={Conditional-flow nerf: Accurate 3d modelling with reliable uncertainty quantification},
  author={Shen, Jianxiong and Agudo, Antonio and Moreno-Noguer, Francesc and Ruiz, Adria},
  booktitle={European Conference on Computer Vision},
  pages={540--557},
  year={2022},
  organization={Springer}
}

@article{carlsson2008local,
  title={On the local behavior of spaces of natural images},
  author={Carlsson, Gunnar and Ishkhanov, Tigran and De Silva, Vin and Zomorodian, Afra},
  journal={International journal of computer vision},
  volume={76},
  number={1},
  pages={1--12},
  year={2008},
  publisher={Springer}
}

@article{fefferman2016testing,
  title={Testing the manifold hypothesis},
  author={Fefferman, Charles and Mitter, Sanjoy and Narayanan, Hariharan},
  journal={Journal of the American Mathematical Society},
  volume={29},
  number={4},
  pages={983--1049},
  year={2016}
}

@inproceedings{gong2019intrinsic,
  title={On the intrinsic dimensionality of image representations},
  author={Gong, Sixue and Boddeti, Vishnu Naresh and Jain, Anil K},
  booktitle={Proceedings of the IEEE/CVF Conference on Computer Vision and Pattern Recognition},
  pages={3987--3996},
  year={2019}
}

@article{pope2021intrinsic,
  title={The intrinsic dimension of images and its impact on learning},
  author={Pope, Phillip and Zhu, Chen and Abdelkader, Ahmed and Goldblum, Micah and Goldstein, Tom},
  journal={arXiv preprint arXiv:2104.08894},
  year={2021}
}

@inproceedings{li2025understanding,
    title={Understanding Representation Dynamics of Diffusion Models via Low-Dimensional Modeling},
    author={Xiao Li and Zekai Zhang and Xiang Li and Siyi Chen and Zhihui Zhu and Peng Wang and Qing Qu},
    booktitle={The Thirty-ninth Annual Conference on Neural Information Processing Systems},
    year={2025},
    url={https://openreview.net/forum?id=BE6QmLdJqY}
}

@article{li2023galaxy,
  title={Galaxy image deconvolution for weak gravitational lensing with unrolled plug-and-play ADMM},
  author={Li, Tianao and Alexander, Emma},
  journal={Monthly Notices of the Royal Astronomical Society: Letters},
  volume={522},
  number={1},
  pages={L31--L35},
  year={2023},
  publisher={Oxford University Press}
}

@article{zbontar2018fastmri,
  title={fastMRI: An open dataset and benchmarks for accelerated MRI},
  author={Zbontar, Jure and Knoll, Florian and Sriram, Anuroop and Murrell, Tullie and Huang, Zhengnan and Muckley, Matthew J and Defazio, Aaron and Stern, Ruben and Johnson, Patricia and Bruno, Mary and others},
  journal={arXiv preprint arXiv:1811.08839},
  year={2018}
}

@article{rudin1992nonlinear,
  title={Nonlinear total variation based noise removal algorithms},
  author={Rudin, Leonid I and Osher, Stanley and Fatemi, Emad},
  journal={Physica D: nonlinear phenomena},
  volume={60},
  number={1-4},
  pages={259--268},
  year={1992},
  publisher={Elsevier}
}

@article{deng2012mnist,
  title={The mnist database of handwritten digit images for machine learning research},
  author={Deng, Li},
  journal={IEEE Signal Processing Magazine},
  volume={29},
  number={6},
  pages={141--142},
  year={2012},
  publisher={IEEE}
}

@inproceedings{choi2020stargan,
  title={Stargan v2: Diverse image synthesis for multiple domains},
  author={Choi, Yunjey and Uh, Youngjung and Yoo, Jaejun and Ha, Jung-Woo},
  booktitle={Proceedings of the IEEE/CVF conference on computer vision and pattern recognition},
  pages={8188--8197},
  year={2020}
}

@article{barco2025blind,
  title={Blind Strong Gravitational Lensing Inversion: Joint Inference of Source and Lens Mass with Score-Based Models},
  author={Barco, Gabriel Missael and Legin, Ronan and Stone, Connor and Hezaveh, Yashar and Perreault-Levasseur, Laurence},
  journal={arXiv preprint arXiv:2511.04792},
  year={2025}
}

@article{adam2022posterior,
  title={Posterior samples of source galaxies in strong gravitational lenses with score-based priors},
  author={Adam, Alexandre and Coogan, Adam and Malkin, Nikolay and Legin, Ronan and Perreault-Levasseur, Laurence and Hezaveh, Yashar and Bengio, Yoshua},
  journal={arXiv preprint arXiv:2211.03812},
  year={2022}
}

@article{karchev2022strong,
  title={Strong-lensing source reconstruction with denoising diffusion restoration models},
  author={Karchev, Konstantin and Montel, Noemi Anau and Coogan, Adam and Weniger, Christoph},
  journal={arXiv preprint arXiv:2211.04365},
  year={2022}
}

@article{zheng2025inversebench,
  title={Inversebench: Benchmarking plug-and-play diffusion priors for inverse problems in physical sciences},
  author={Zheng, Hongkai and Chu, Wenda and Zhang, Bingliang and Wu, Zihui and Wang, Austin and Feng, Berthy T and Zou, Caifeng and Sun, Yu and Kovachki, Nikola and Ross, Zachary E and others},
  journal={arXiv preprint arXiv:2503.11043},
  year={2025}
}

@inproceedings{
    erbach2025solving,
    title={Solving Inverse Problems with {FLAIR}},
    author={Julius Erbach and Dominik Narnhofer and Andreas Robert Dombos and Bernt Schiele and Jan Eric Lenssen and Konrad Schindler},
    booktitle={The Thirty-ninth Annual Conference on Neural Information Processing Systems},
    year={2025},
    url={https://openreview.net/forum?id=w9xETx7HT1}
}

\ifpeerreview \else


\begin{IEEEbiography}[{\includegraphics[width=1in,height=1.25in,clip,keepaspectratio]{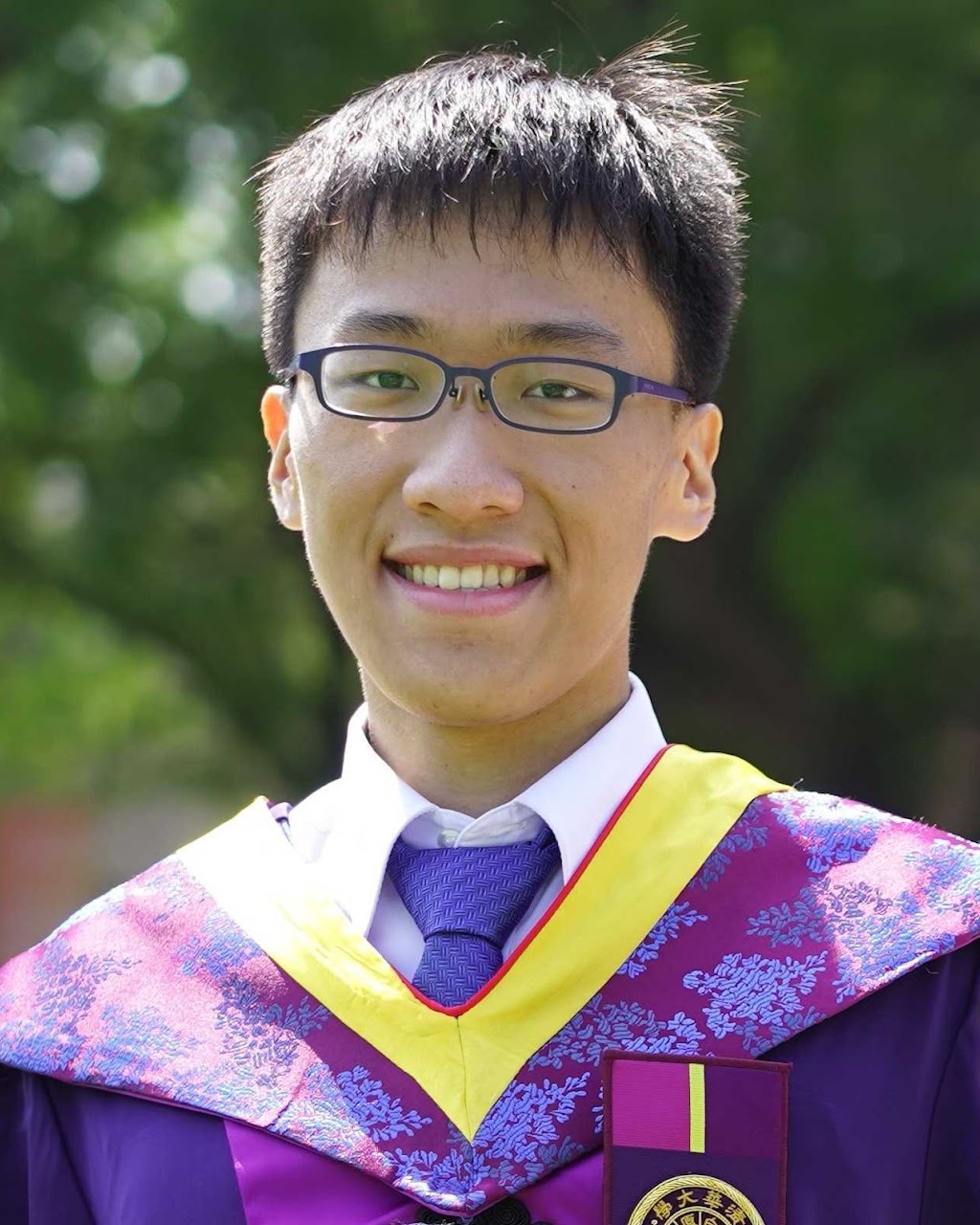}}]{Tianao Li} (Student Member, IEEE) received the B.Eng. degree in electronic engineering with a minor in Astronomy from Tsinghua University, Beijing, China, in 2019, and the M.S. degree in computer science from Northwestern University, Evanston, IL, USA, in 2026.
He is currently a Ph.D. candidate in the Department of Computer Science at Northwestern University, Evanston, IL, USA.
His research focuses on how generative priors, scene representations, and physics-based forward modeling can advance imaging — turning ill-posed measurements into faithful reconstructions of the world.
\end{IEEEbiography}

\begin{IEEEbiography}[{\includegraphics[width=1in,height=1.25in,clip,keepaspectratio]{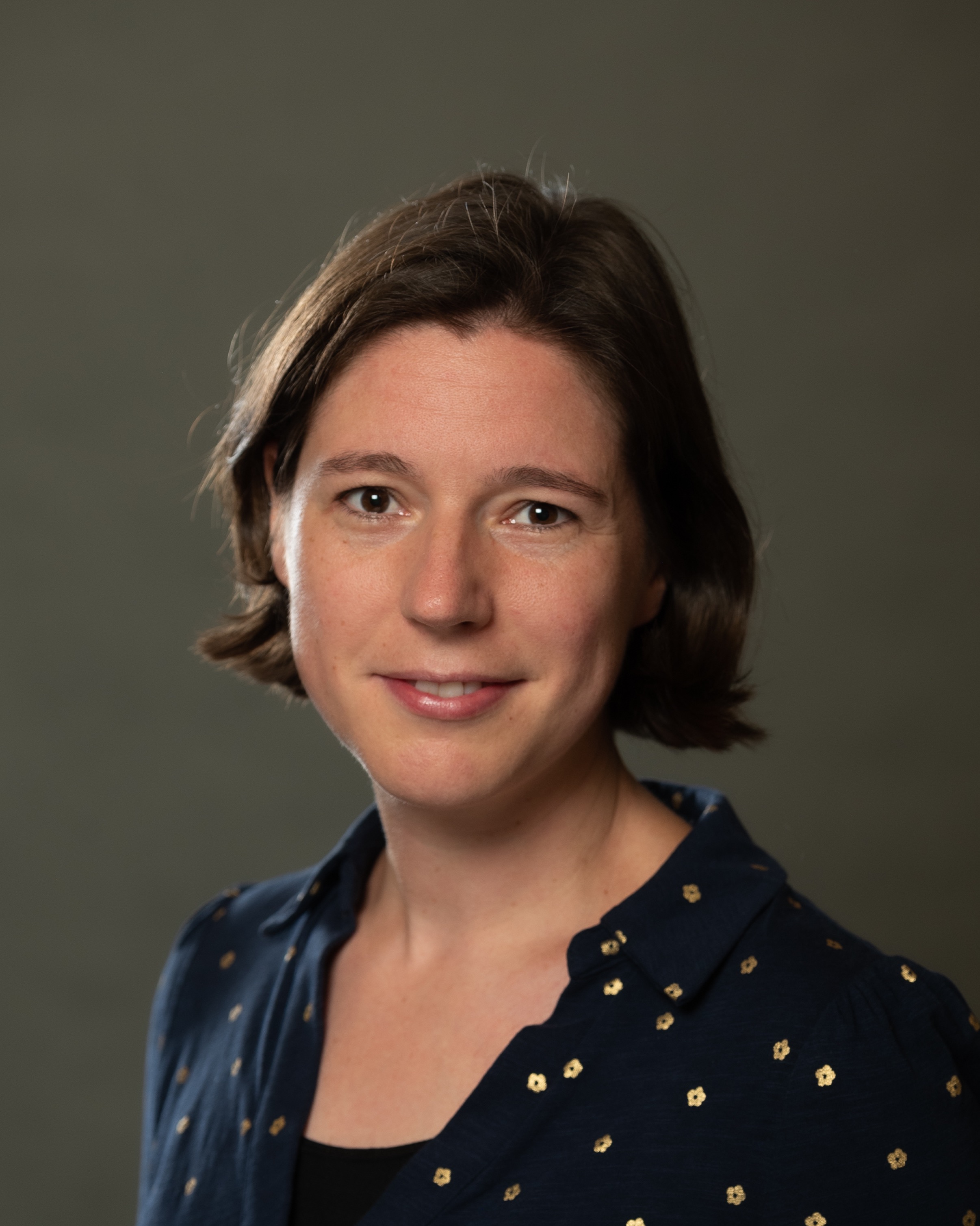}}]{Tjitske Starkenburg} received a double B.Sc. in Astronomy and Mathematics, and a M.Sc. and PhD in Astronomy from the University of Groningen, the Netherlands.
She has been a Flatiron Research Fellow at the Center for Computational Astrophysics of the Flatiron Institute in New York, New York, USA, and a CIERA Fellow at Northwestern University, Evanston, IL, USA. Starkenburg is currently an Assistant Research Professor with the Department of Physics and Astronomy and the Center for Interdisciplinary Exploration and Research in Astrophysics (CIERA) at Northwestern University and the NSF-Simons AI Institute for the Sky (SkAI). Her research focuses on the formation and evolution of galaxies in the universe, computational astrophysics and inference.
\end{IEEEbiography}

\begin{IEEEbiography}[{\includegraphics[width=1in,height=1.25in,clip,keepaspectratio]{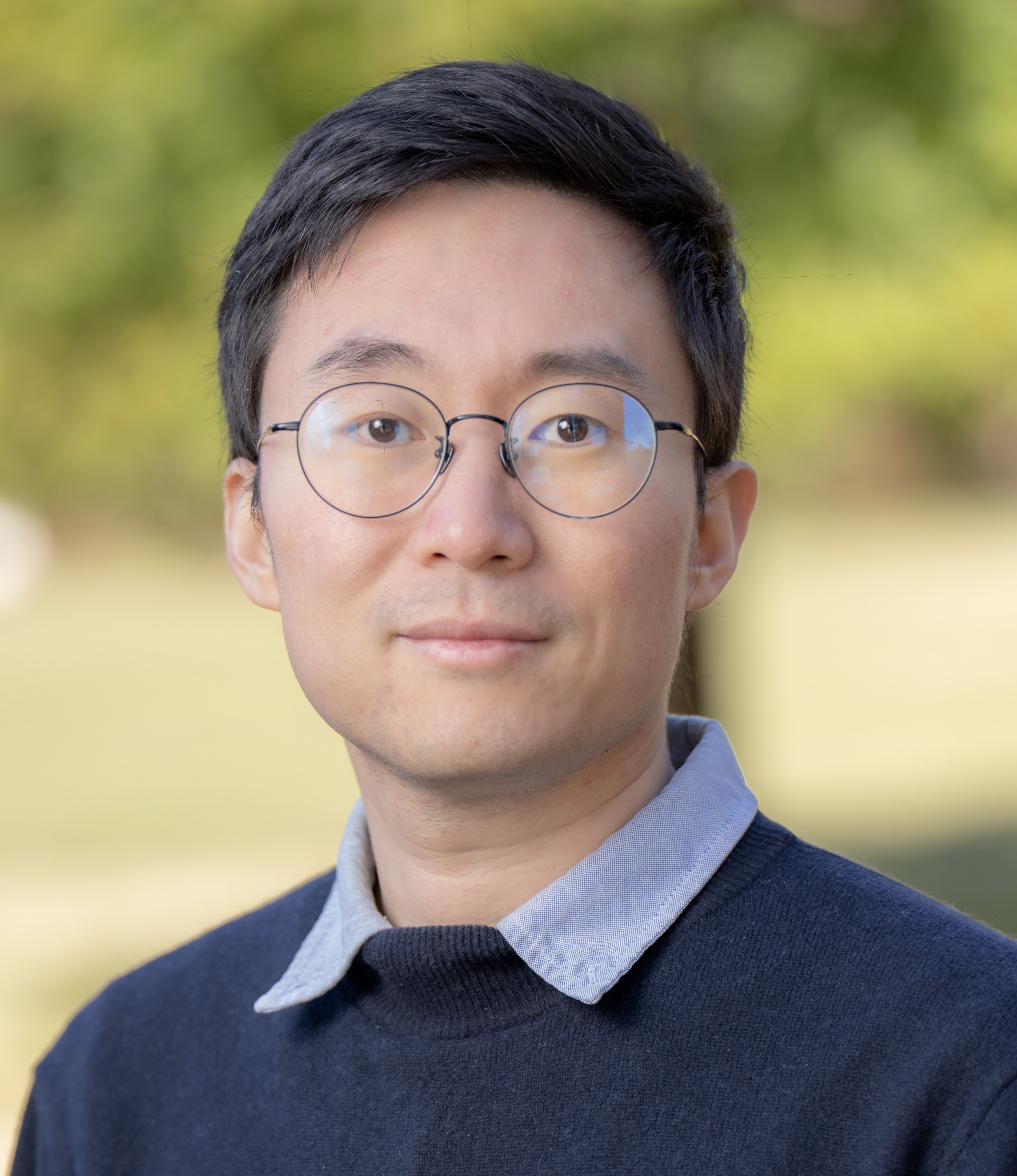}}]
{Yu Sun} (Member, IEEE) received the B.Eng. degree in electronics and information from Sichuan University, Chengdu, China, in 2015, and the Ph.D. degree in computer science from Washington University in St. Louis, St. Louis, MO, USA, in 2022.
He is currently an Assistant Professor with the Department of Electrical and Computer Engineering, Johns Hopkins University, Baltimore, MD, USA. Prior to joining Johns Hopkins, he was a Computing, Data \& Society Fellow with the Department of Computing and Mathematical Sciences, California Institute of Technology, Pasadena, CA, USA.
His research focuses on explainable and reliable AI algorithms for scientific imaging and computer vision. His doctoral dissertation received the Turner Dissertation Award in Computer Science at Washington University in St. Louis. He is a recipient of the NSF CAREER Award and a member of the IEEE Signal Processing Society’s Computational Imaging Technical Committee.
\end{IEEEbiography}

\begin{IEEEbiography}[{\includegraphics[width=1in,height=1.25in,clip,keepaspectratio]{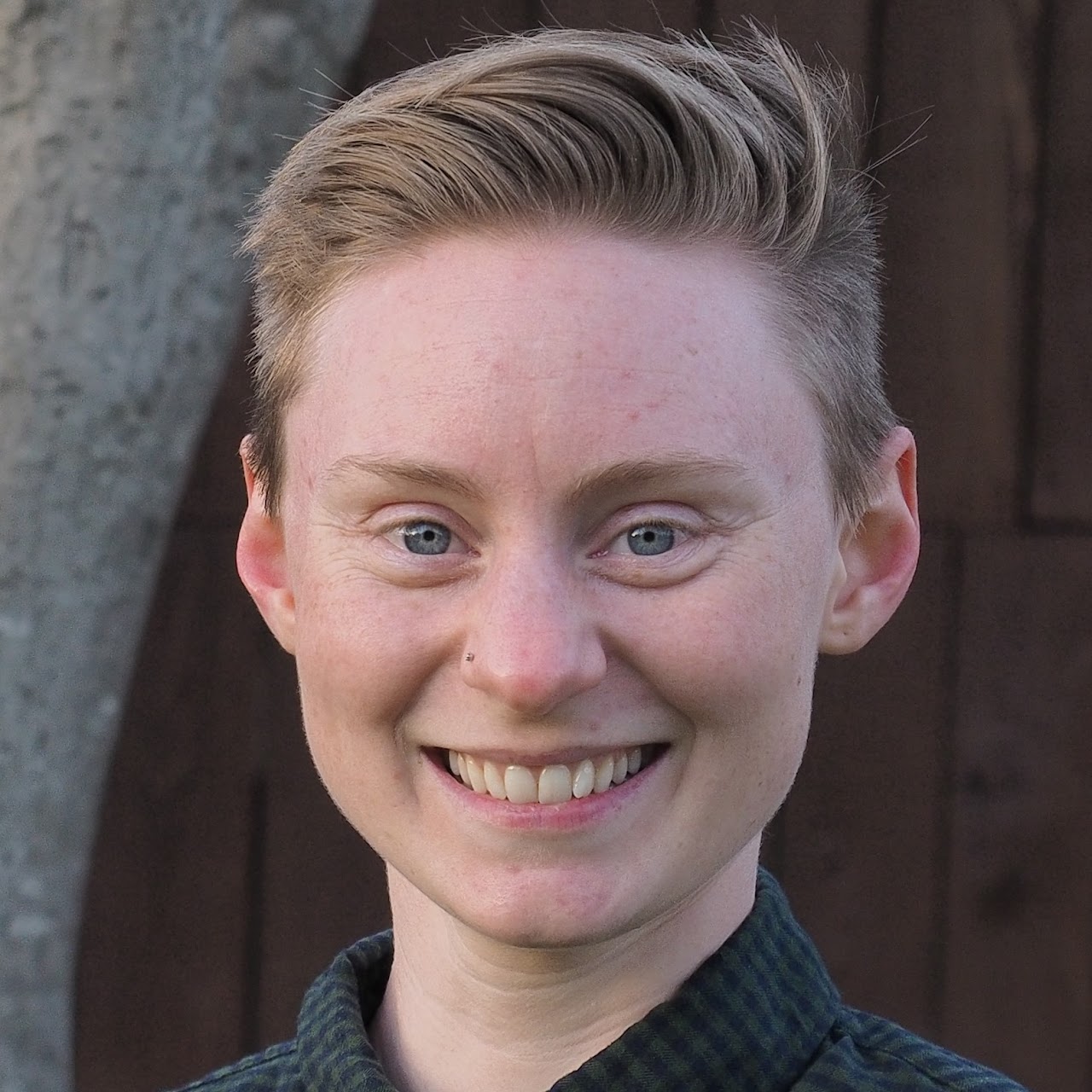}}]
{Emma Alexander} (Member, IEEE) received the B.S. degree in physics and in computer science from Yale University, New Haven, CT, USA, in 2013, and the Ph.D. degree in computer science from Harvard University in Cambridge, MA, USA, in 2019.
She is currently an Assistant Professor with the Department of Computer Science, Northwestern University, Evanston, IL, USA. Prior to joining Northwestern, she was a postdoctoral researcher at the University of California Berkeley, Berkeley, CA, USA.
Her research focuses on bio-inspired computational imaging. She is a recipient of the Fulbright Specialist Award and a member of the IEEE Signal Processing Society’s Computational Imaging Technical Committee.
\end{IEEEbiography}




\fi

\clearpage


\newcounter{si}
\setcounter{si}{0} 
\renewcommand\thesection{S\arabic{si}}

\newcounter{fi}
\setcounter{fi}{0} 
\renewcommand{\thefigure}{S\arabic{fi}}

\newcounter{ei}
\setcounter{ei}{0} 
\renewcommand{\theequation}{S\arabic{ei}}

\newcounter{ti}
\setcounter{ti}{0} 
\renewcommand{\thetable}{S\arabic{ti}}

\twocolumn[{%
  \centering\huge ShuffleFlow: Scalable Posterior Inference for Bayesian Inverse Imaging \\[0.4em]
  \LARGE Supplementary Material
  \vspace{1.5em}\par}
]

\stepcounter{si}
\section{Inverse problem setup}
\label{supp:inverse_problems}

In this section, we describe setups of the inverse problems discussed in Sec.~\ref{sec:results}.  

\subsection{Toy example}
\label{supp:toy_example}

In our toy example, we use digit ``3" images from the MNIST dataset~\cite{deng2012mnist} (resized to 32$\times$32 and normalized to [-1,1]). We calculate a Gaussian prior $\mathcal{N}(\mathbf{m},\mathbf{\Lambda})$ from all ``3" images in the dataset, and use a diagonal covariance matrix for simplicity. 
The forward model is 
\stepcounter{ei}
\begin{equation}
    \mathbf{y=Ax+n},
\end{equation}
where $\mathbf{y}$ is the observed image, $\mathbf{A}$ is a 512$\times$1024 random Gaussian matrix, $\mathbf{x}$ is the clean image, and $\mathbf{n}$ is i.i.d. Gaussian noise with $\sigma=0.1$.
This gives us a Gaussian data likelihood $\mathcal{N}(\mathbf{Ax}, \sigma^2\mathbf{I})$.
In this case, we are able to analytically compute the Gaussian posterior $\mathcal{N}(\boldsymbol{\mu},\mathbf{\Sigma})$:
\stepcounter{ei}
\begin{equation}
    \begin{aligned}
        \boldsymbol{\mu} &= \left( \frac{1}{\sigma^2} \mathbf{A}^{T}\mathbf{A} + \mathbf{\Lambda}^{-1} \right)^{-1}\left( \frac{1}{\sigma^2} \mathbf{A}^{T}\mathbf{y} + \mathbf{\Lambda}^{-1}\mathbf{m} \right), \\
        \mathbf{\Sigma} &= \left( \frac{1}{\sigma^2} \mathbf{A}^{T}\mathbf{A} + \mathbf{\Lambda}^{-1} \right)^{-1}.
    \end{aligned}
\end{equation}

\begin{figure}[b]
    \centering
    \includegraphics[width=\linewidth]{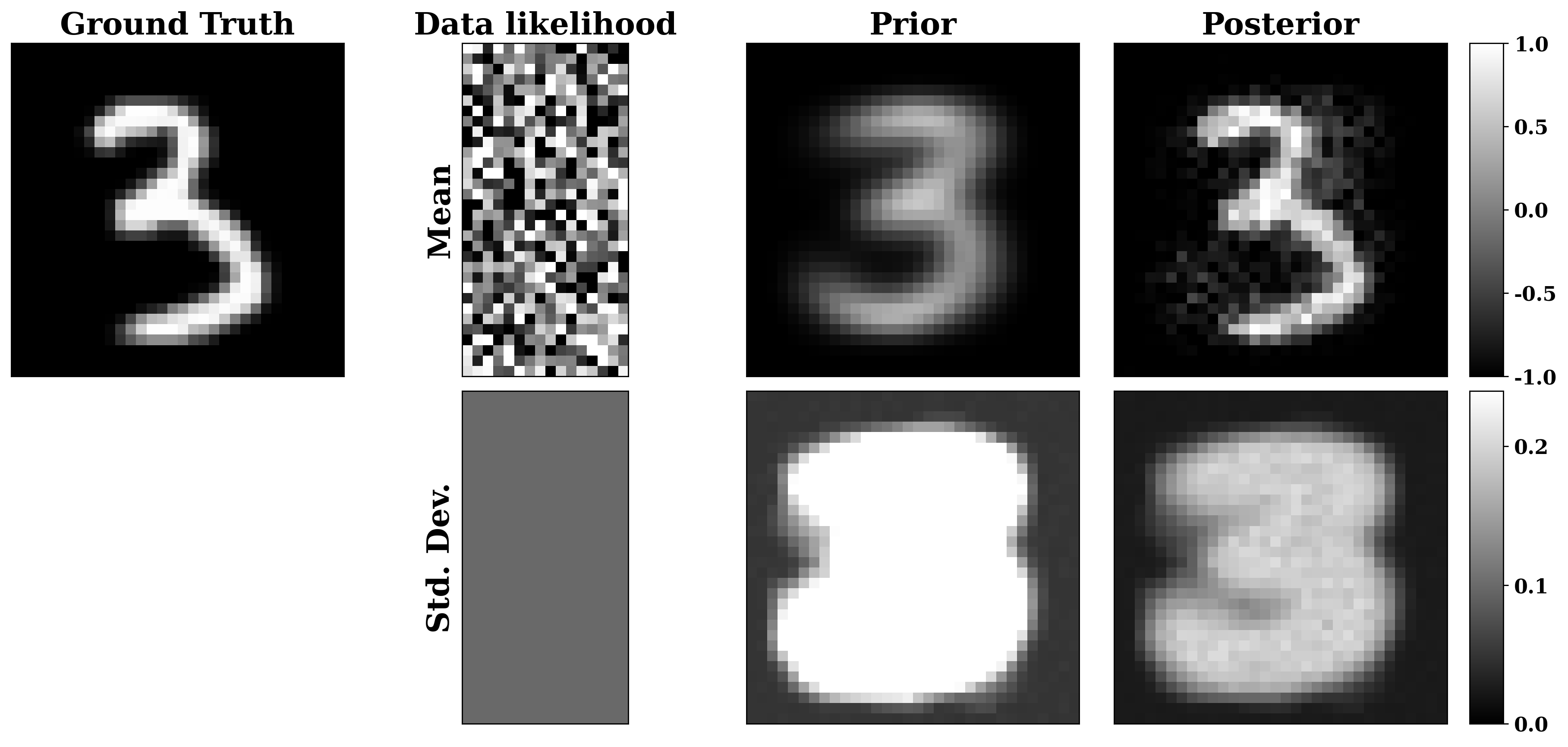}
    \caption{
        {\bf Inverse problem used in toy example.}
    }
    \label{fig:toy_setup}
\end{figure}

We visualize the ground truth image, data likelihood, prior, and posterior used in Sec.~\ref{sec:toy_example} in Fig.~\ref{fig:toy_setup}.
Note that we lose half of the image dimensions in the measurement $\mathbf{y}$ because we have a 512$\times$1024 forward matrix $\mathbf{A}$.
Since the digits are centered in the images, our Gaussian prior has high standard deviations in the center and very low standard deviations in the background.
This leads to stronger regularizations in the background compared to the image center.
Therefore, in the posterior mean image, we see very little noise in the background but very obvious noise around the image center.

\stepcounter{fi}

\subsection{Motion deblurring}
\label{supp:deblur}

\stepcounter{fi}
\begin{figure*}
    \centering
    \includegraphics[width=\linewidth]{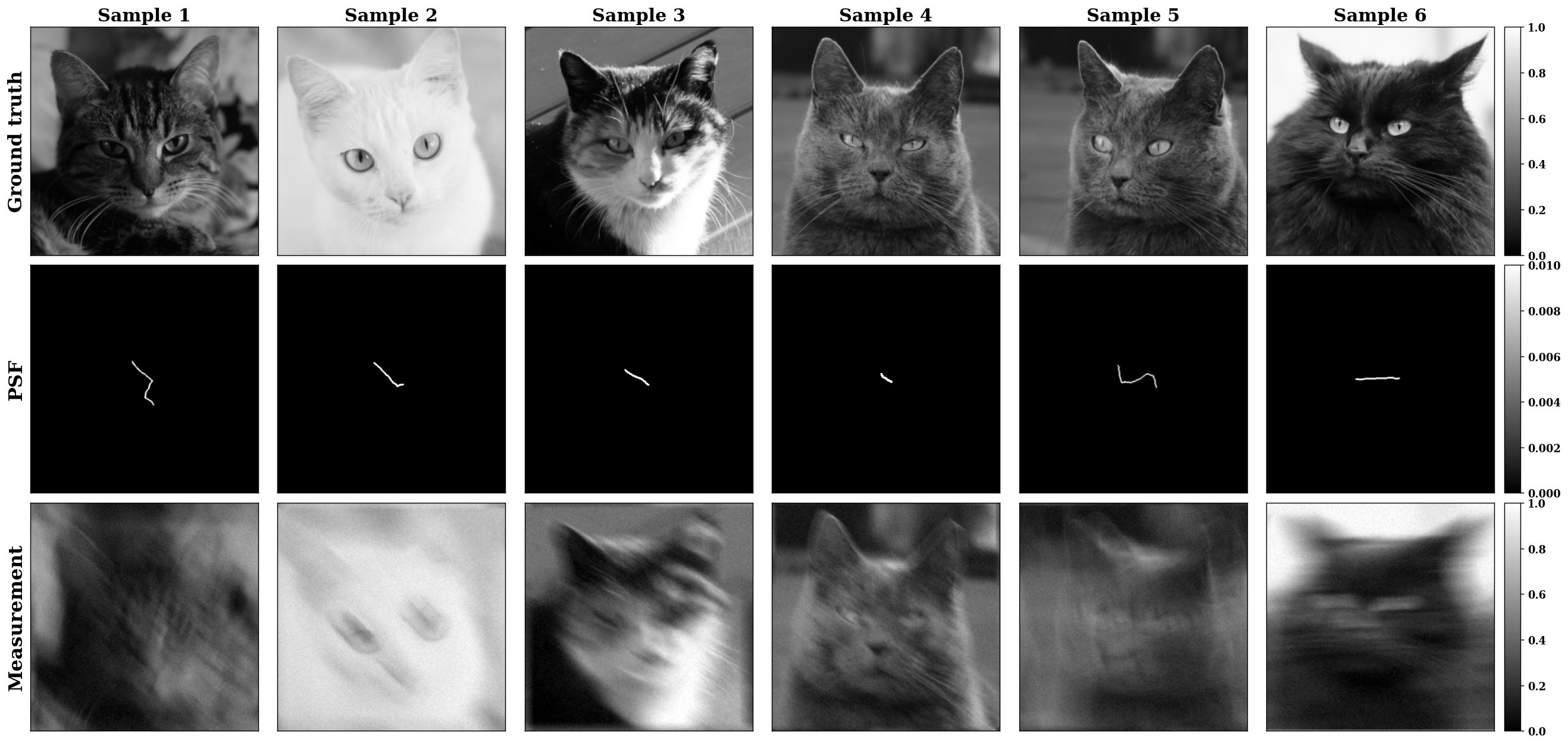}
    \caption{
        {\bf Samples of the motion deblurring dataset.}
    }
    \label{fig:deblur_setup}
\end{figure*}

In our motion deblurring experiment, we use data from the AFHQ-v2 dataset~\cite{choi2020stargan}.
All images are resized to 256$\times$256 and normalized to [0,1].
We train our score model on all 12,902 samples in the training set.
The forward model is
\stepcounter{ei}
\begin{equation}
    \mathbf{y = Hx + n},
    \label{eq:deblur}
\end{equation}
where $\mathbf{y}$ is the observed image, $\mathbf{H}$ is the convolution matrix of the blur kernel, $\mathbf{x}$ is the clean image, and $\mathbf{n}$ is i.i.d. Gaussian noise with $\sigma=0.02$.
We generate motion blur kernels\footnote{\href{https://github.com/LeviBorodenko/motionblur}{https://github.com/LeviBorodenko/motionblur}} with kernel size 61$\times$61 and intensity of 0.5.
We show several samples of ground truth images (row 1), motion blur kernels (row 2), and corresponding measurements (row 3) in Fig.~\ref{fig:deblur_setup}.

\subsection{Compressed sensing MRI}
\label{supp:mri}

\stepcounter{fi}
\begin{figure*}
    \centering
    \includegraphics[width=\linewidth]{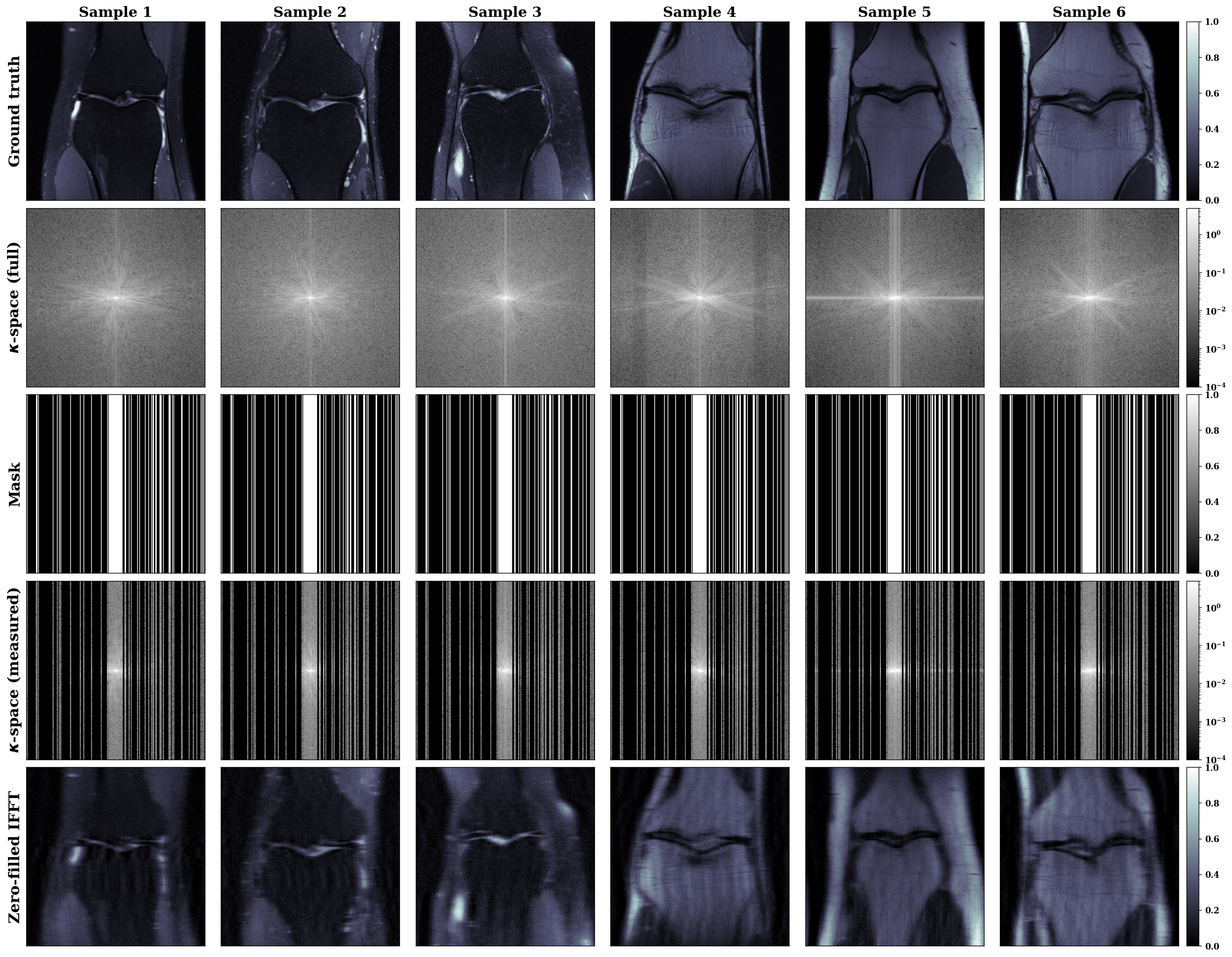}
    \caption{
        {\bf Samples of the MRI dataset.}
    }
    \label{fig:mri_setup}
\end{figure*}

In our MRI experiment, we use data from the fastMRI single coil knee dataset~\cite{zbontar2018fastmri}.
All images are resized to 256$\times$256 and normalized to $[0,1]$.
We train our score model on all 20,443 samples in the training set.
The forward model can be expressed as: 
\stepcounter{ei}
\begin{equation}
    \mathbf{y = M\mathcal{F}(x) + n},
    \label{eq:mri}
\end{equation}
where $\mathbf{y}$ is the measured k-space components, $\mathbf{M}$ is the sampling matrix, $\mathcal{F}(\cdot)$ is the 2D Fourier transform, $\mathbf{x}$ is the clean image, and $\mathbf{n}$ is i.i.d. Gaussian noise.
We simulate Cartesian sampling masks using~\cite{zheng2025inversebench}'s implementation with an acceleration rate of 4.0.
Specifically, we fully sample 8.0\% of k-space components centered around DC and sample equidistance lines in the rest area with a spacing that achieves the desired acceleration rate.
We add i.i.d. Gaussian noise with a standard deviation of 0.02.
We show several samples of the ground truth image (row 1), full k-space components (row 2), cartesian sampling mask (row 3), measured k-space (row 4), and zero-filled IFFT reconstruction (row 5) in Fig.~\ref{fig:mri_setup}.

\subsection{Fourier phase retrieval}
\label{supp:pr}

\stepcounter{fi}
\begin{figure*}
    \centering
    \includegraphics[width=\linewidth]{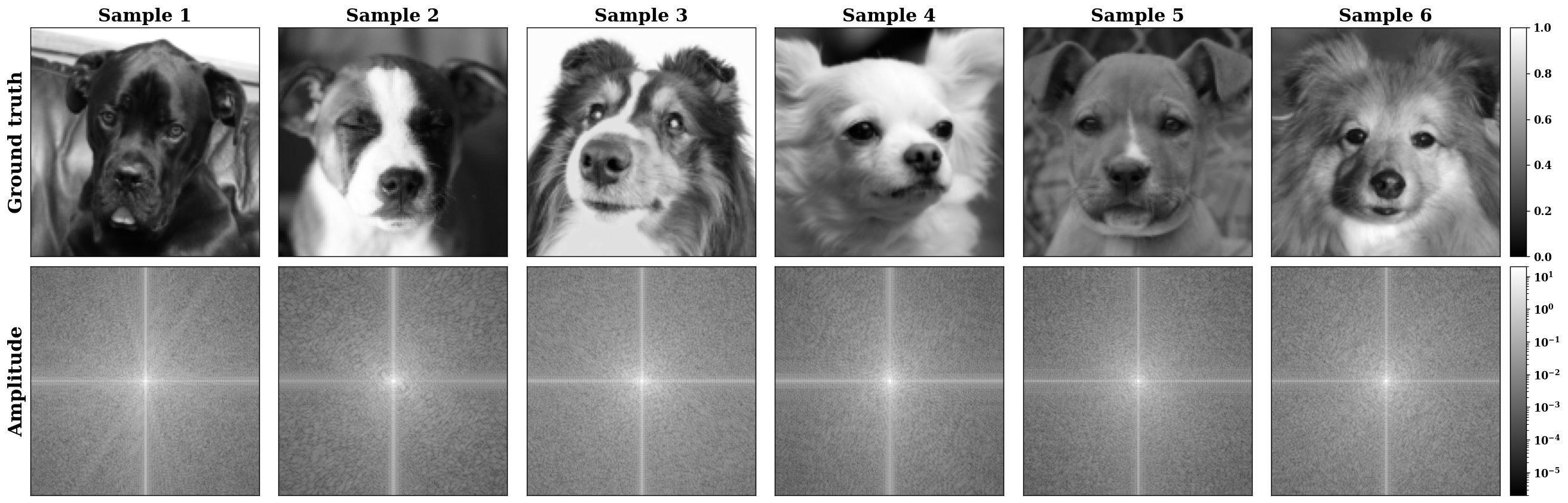}
    \caption{
        {\bf Samples of the Fourier phase retrieval dataset.}
    }
    \label{fig:fpr_setup}
\end{figure*}

In our Fourier phase retrieval experiment, we also use data from the AFHQ-v2 dataset~\cite{choi2020stargan} but instead resize the images to 128$\times$128.
The forward model can be expressed as
\stepcounter{ei}
\begin{equation}
    \mathbf{y = \left\vert \mathcal{F}(Px) \right\vert + n},
    \label{eq:fpr}
\end{equation}
where $\mathbf{y}$ is the measured Fourier space amplitude, $\mathcal{F}(\cdot)$ is the 2D Fourier transform, $|\cdot|$ indicates the amplitude calculation (removing phase information), $\mathbf{P}$ is the zero-padding operator, $\mathbf{x}$ is the clean image, and $\mathbf{n}$ is i.i.d. Gaussian noise with $\sigma=0.01$.
To reduce the ill-posedness of the problem, we apply oversampling by zero-padding the image to 2$\times$ of its original size to increase the number of Fourier amplitude measurements, which is a common practice for this problem~\cite{wu2024principled}.

We show several samples of ground truth images (row 1) and Fourier amplitude measurements (row 2) in Fig.~\ref{fig:fpr_setup}.

\clearpage

\stepcounter{ti}
\begin{table*}[h]
    \footnotesize
    \setlength{\tabcolsep}{3pt} 
    \centering
    \caption{
        {\bf Quantitative results for linear inverse problems with TV prior.}
    }
    \label{tab:linear_tv}
    \begin{tabular}{@{}lcccccccccccc@{}}
        \toprule
        \multirow{2}{*}{\bf Method}
        & \multicolumn{6}{c}{\bf Motion Deblurring}
        & \multicolumn{6}{c}{\bf Compressed Sensing MRI} \\
        \cmidrule(lr){2-7} \cmidrule(lr){8-13} 
        & \bf PSNR $\mathbf{\uparrow}$ & \bf SSIM $\mathbf{\uparrow}$ & \bf LPIPS $\mathbf{\downarrow}$ & \bf NLL $\mathbf{\downarrow}$ & \bf ECE $\mathbf{\downarrow}$ & \bf OODR $\mathbf{\downarrow}$
        & \bf PSNR $\mathbf{\uparrow}$ & \bf SSIM $\mathbf{\uparrow}$ & \bf LPIPS $\mathbf{\downarrow}$ & \bf NLL $\mathbf{\downarrow}$ & \bf ECE $\mathbf{\downarrow}$ & \bf OODR $\mathbf{\downarrow}$ \\
        \midrule
        DPI~\cite{sun2021deep}
        & 24.11 & 0.453 & 0.470 & -1.31 & 0.069 & 0.09\% 
        & 29.02 & 0.736 & 0.312 & -1.25 & 0.097 & 9.52\% \\
        CF-NeRF~\cite{shen2022conditional}
        & 25.58 & 0.548 & 0.418 & -1.42 & 0.091 & 0.09\% 
        & 29.33 & 0.743 & 0.309 & -1.00 & 0.111 & 11.78\% \\
        Ours w/ patch
        & 24.97 & 0.521 & 0.460 & -1.31 & 0.141 & 0.13\% 
        & 29.07 & 0.733 & 0.307 & -1.48 & 0.079 & 7.39\% \\
        Ours
        & 27.01 & 0.669 & 0.415 & -1.57 & 0.105 & 1.73\% 
        & 30.32 & 0.775 & 0.297 & -1.58 & 0.073 & 7.92\% \\
        \midrule
        Zero-filled IFFT
        & N/A & N/A & N/A & N/A & N/A & N/A
        & 26.17 & 0.608 & 0.383 & N/A & N/A & N/A \\
        Wiener
        & 22.70 & 0.378 & 0.542 & N/A & N/A & N/A
        & N/A & N/A & N/A & N/A & N/A & N/A \\
        RML
        & 27.46 & 0.697 & 0.436 & N/A & N/A & N/A 
        & 30.13 & 0.764 & 0.305 & N/A & N/A & N/A \\
        ALD
        & 26.92 & 0.625 & 0.405 & -0.88 & 0.315 & 0.00\%
        & 29.44 & 0.759 & 0.323 & -0.72 & 0.126 & 14.04\% \\
        \bottomrule
    \end{tabular}
\end{table*}

\stepcounter{fi}
\begin{figure*}[h]
    \centering
    \includegraphics[width=\linewidth]{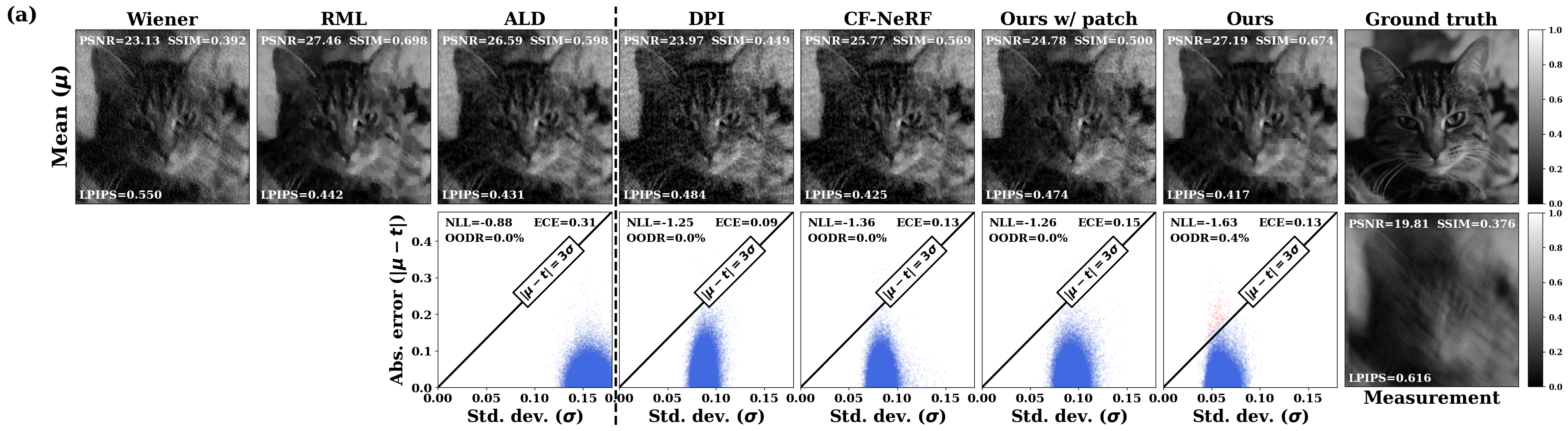}
    \includegraphics[width=\linewidth]{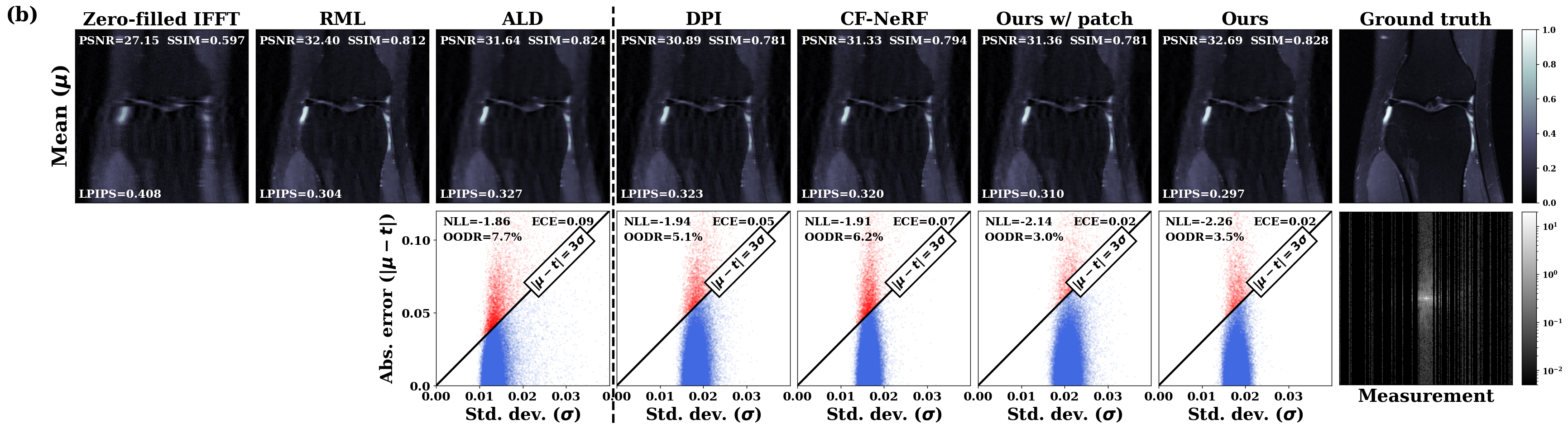}
    \caption{
        {\bf Visual samples on linear inverse problems with TV prior: motion deblurring (a) and compressed sensing MRI (b).}
        We show the mean image computed from 128 posterior samples in row one and the scatter plot of absolute pixel errors vs. standard deviations in row two.
    }
    \label{fig:linear_tv}
\end{figure*}

\stepcounter{si}
\section{Additional results}
\label{supp:results}

\subsection{Linear inverse problems with TV prior}

We show motion deblurring and compressed sensing MRI results with TV prior in Tab.~\ref{tab:linear_tv} and Fig.~\ref{fig:linear_tv}.
We include the following methods as baselines: regularized maximum likelihood (RML), annealed Langevin dynamics (ALD), Wiener deconvolution (for motion deblurring), and zero-filled IFFT (for MRI).
We observe that our method consistently outperforms existing VI methods.

\subsection{PnP-DM results}

\begin{figure*}[h]
    \centering
    \includegraphics[width=\linewidth]{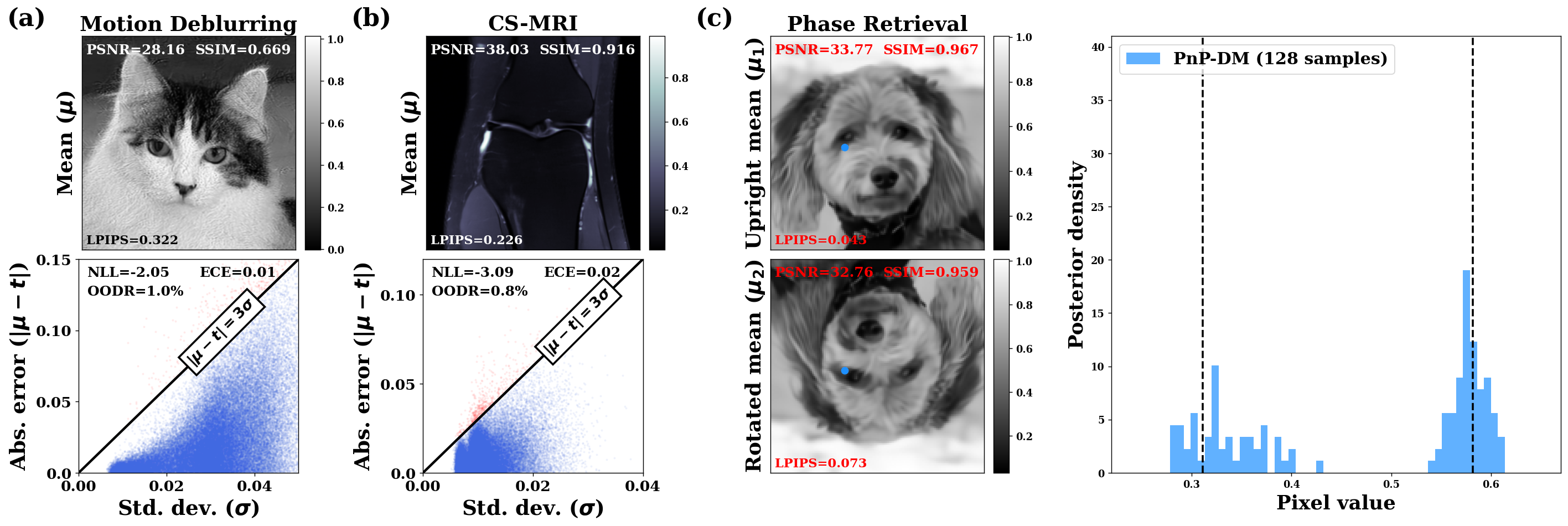}
    \caption{
        {\bf Illustration of PnP-DM~\cite{wu2024principled} results on motion deblurring (a), compressed sensing MRI (b), and Fourier phase retrieval (c).}
        In Fourier phase retrieval, we show the pixel histogram from 128 posterior samples of the same pixel selected in Fig.~\ref{fig:fpr}.
    }
    \label{fig:pnp-dm}
\end{figure*}

We show visual samples of PnP-DM~\cite{wu2024principled} on motion deblurring, compressed sensing MRI, and Fourier phase retrieval in Fig.~\ref{fig:pnp-dm}.
The test samples are the same as the ones shown in the main paper.

\stepcounter{fi}

\subsection{Additional results on Fourier phase retrieval}

\begin{figure*}[h]
    \centering
    \includegraphics[width=0.97\columnwidth]{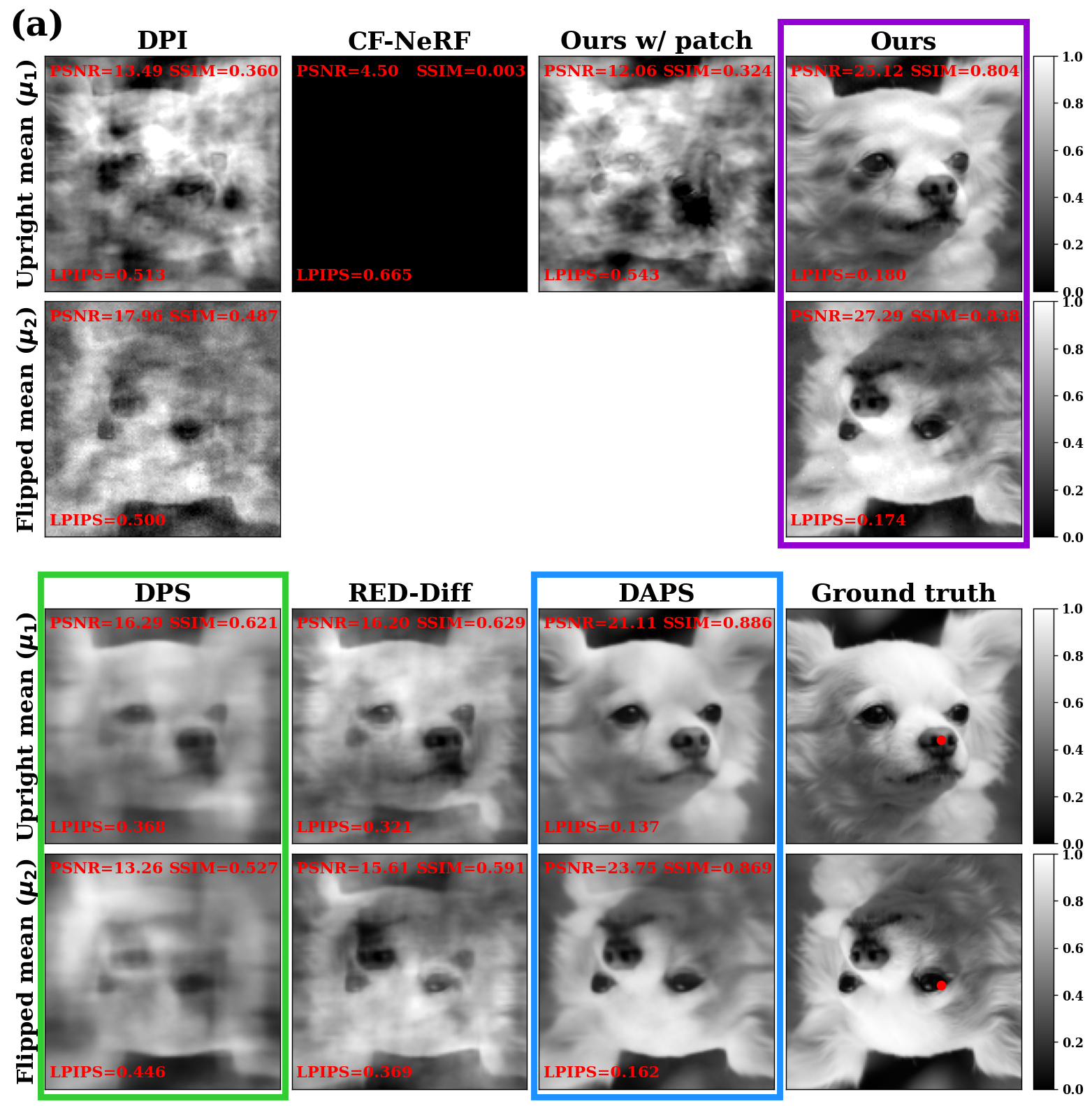}
    \hfill
    \includegraphics[width=1.02\columnwidth]{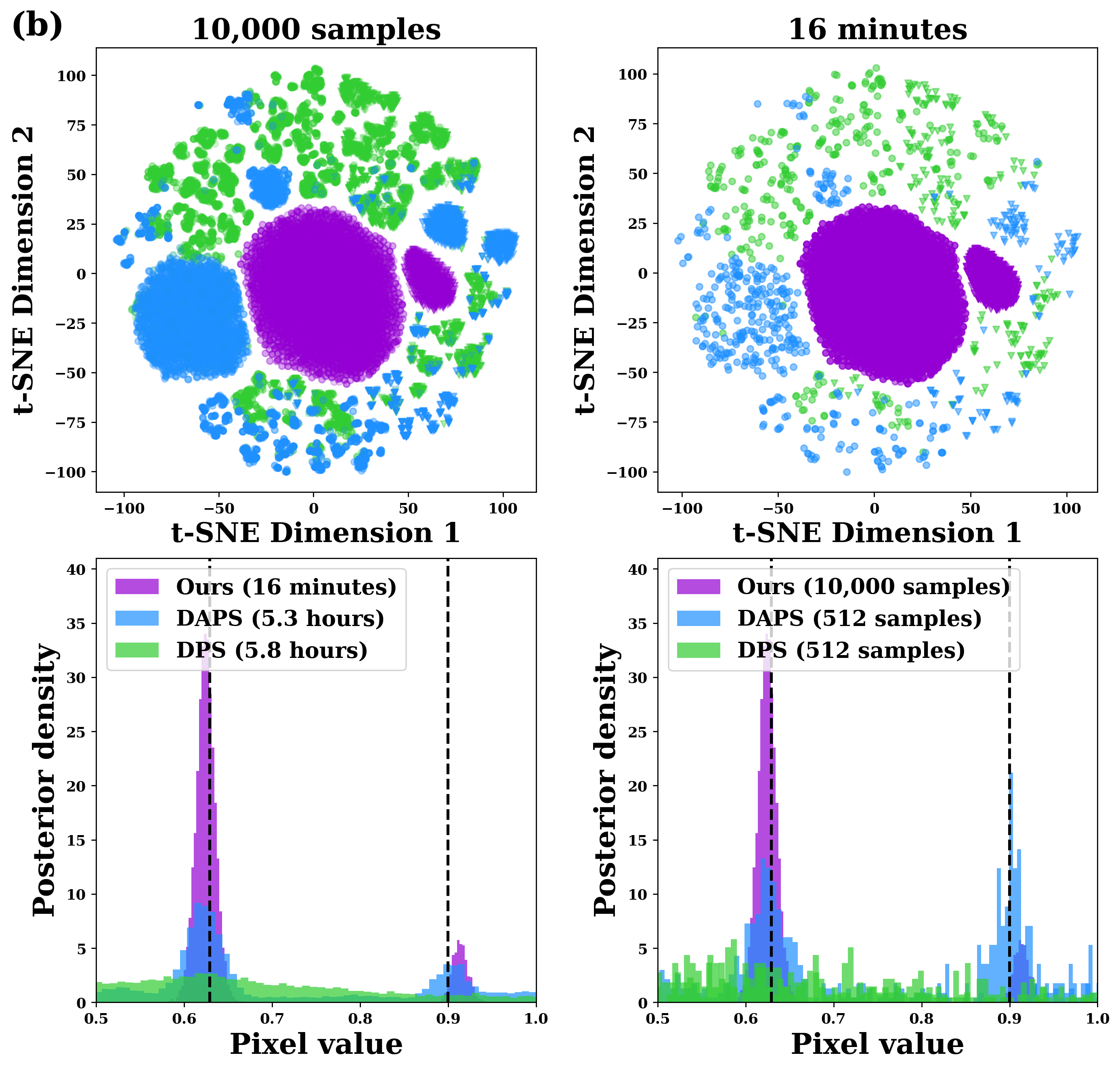}
    \caption{
        {\bf Additional results on nonlinear Fourier phase retrieval.}
        While diffusion samplers produce high image quality, they are unable to show a clear bimodal posterior in a short computation time (see t-SNE visualization and pixel histogram), while our method can.
    }
    \label{fig:fpr_supp}
\end{figure*}

We provide visual results on another test sample for Fourier phase retrieval in Fig.~\ref{fig:fpr_supp}. 
We also picked a pixel (marked with red dots) and show the t-SNE visualization and pixel histograms for a 10,000-sample setting and a short computation time setting.

\begin{figure*}[h]
    \centering
    \includegraphics[width=\linewidth]{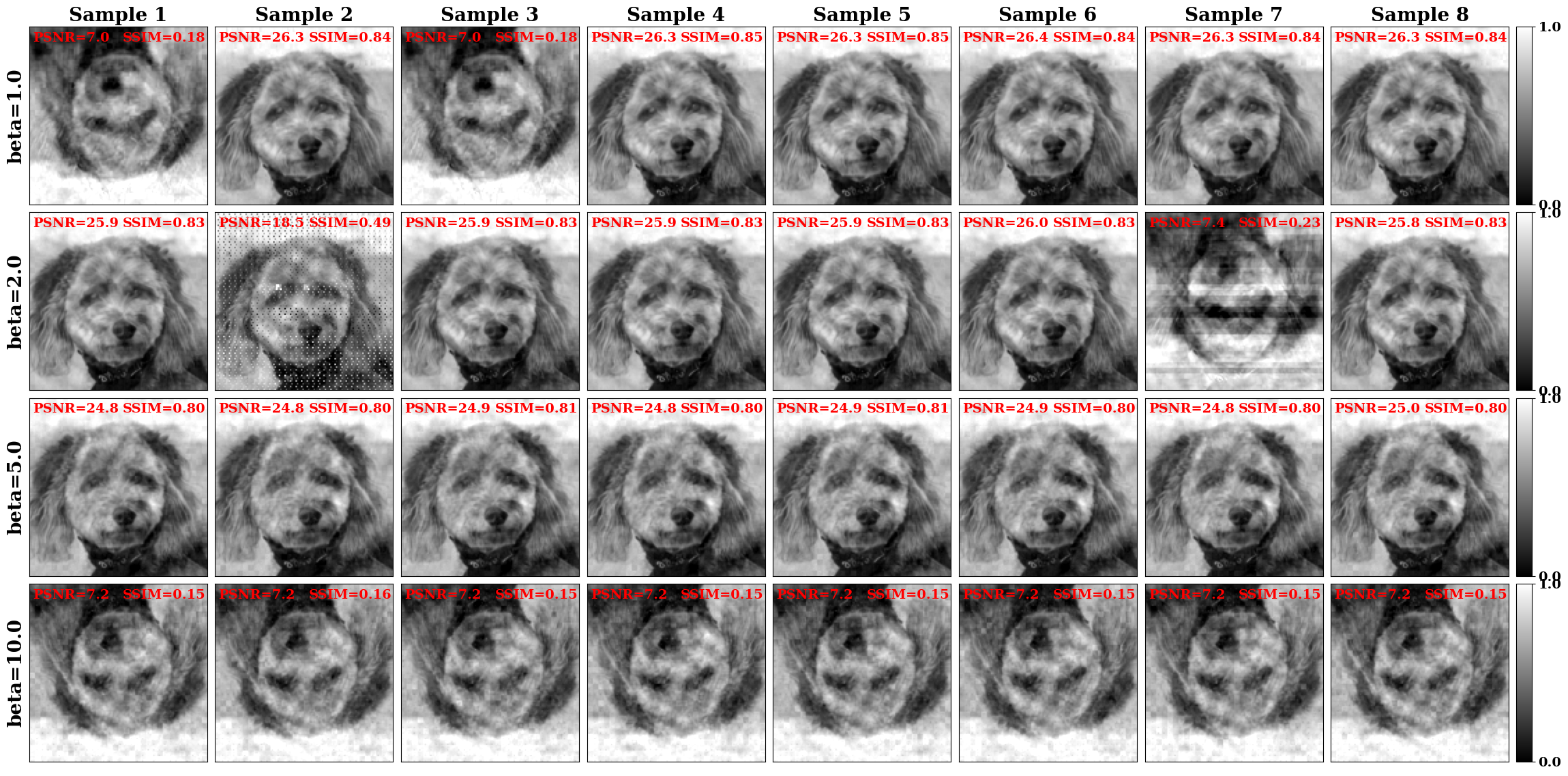}
    \caption{
        {\bf Fourier phase retrieval results with larger entropy weights.}
        We observe artifacts ($\beta=2.0$) and mode collapse with larger entropy weights ($\beta=5.0, 10.0$).
    }
    \label{fig:fpr_beta}
\end{figure*}

In Sec.~\ref{sec:fpr}, we mentioned that VI methods generally underestimate uncertainty in phase retrieval. We show that while this could be theoretically addressed by increasing the entropy weight $\beta$ during training, larger $\beta$ leads to unstable training and mode collapse (see Fig.~\ref{fig:fpr_beta}). This overconfidence/training instability problem should be addressed by future work.

\stepcounter{fi}

\stepcounter{fi}

\subsection{Error maps and individual samples}

\begin{figure*}[h]
    \centering
    \includegraphics[width=\linewidth]{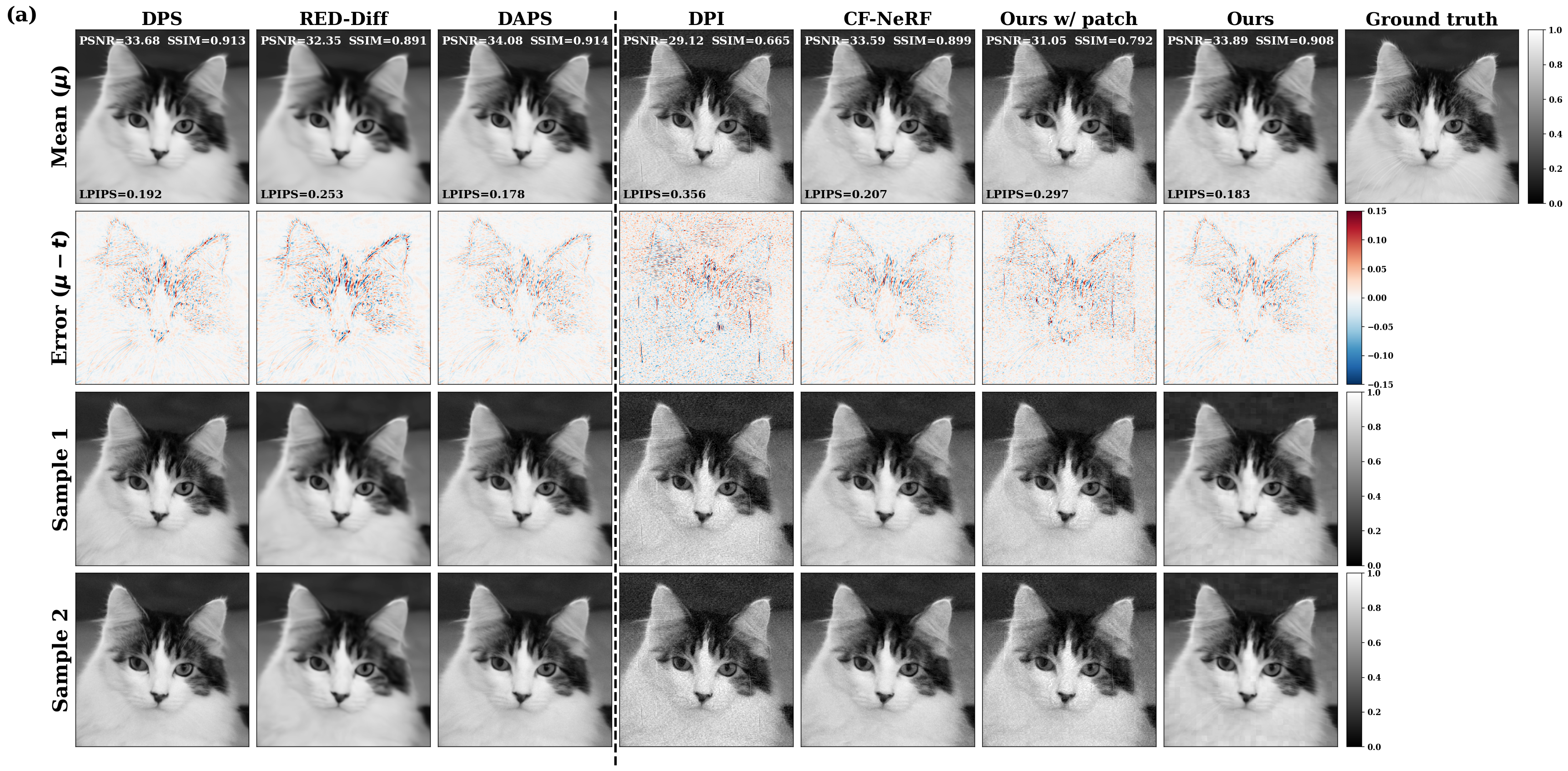}
    \includegraphics[width=\linewidth]{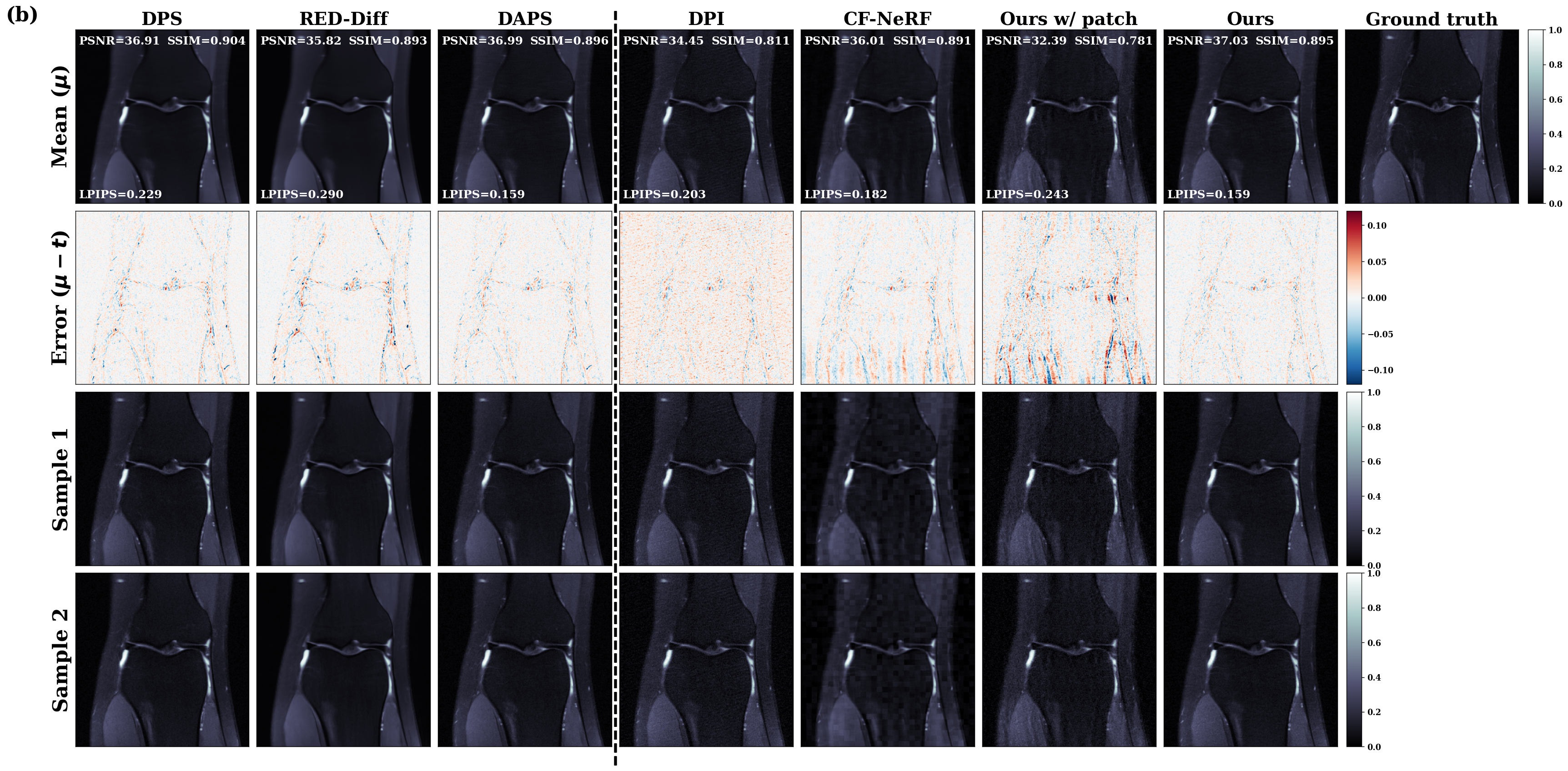}
    
    \caption{
        {\bf Error maps and individual samples for motion deblurring (a) and compressed sensing MRI (b).}
    }
    \label{fig:linear_supp}
\end{figure*}

In Fig.~\ref{fig:linear_supp}, we show error maps and individual samples for Fig.~\ref{fig:linear_score}.

\stepcounter{fi}

\clearpage

\stepcounter{si}
\section{Ablation studies}
\label{supp:ablations}

\subsection{Flow dimensionality}

\begin{figure*}
    \centering
    \includegraphics[width=0.55\linewidth]{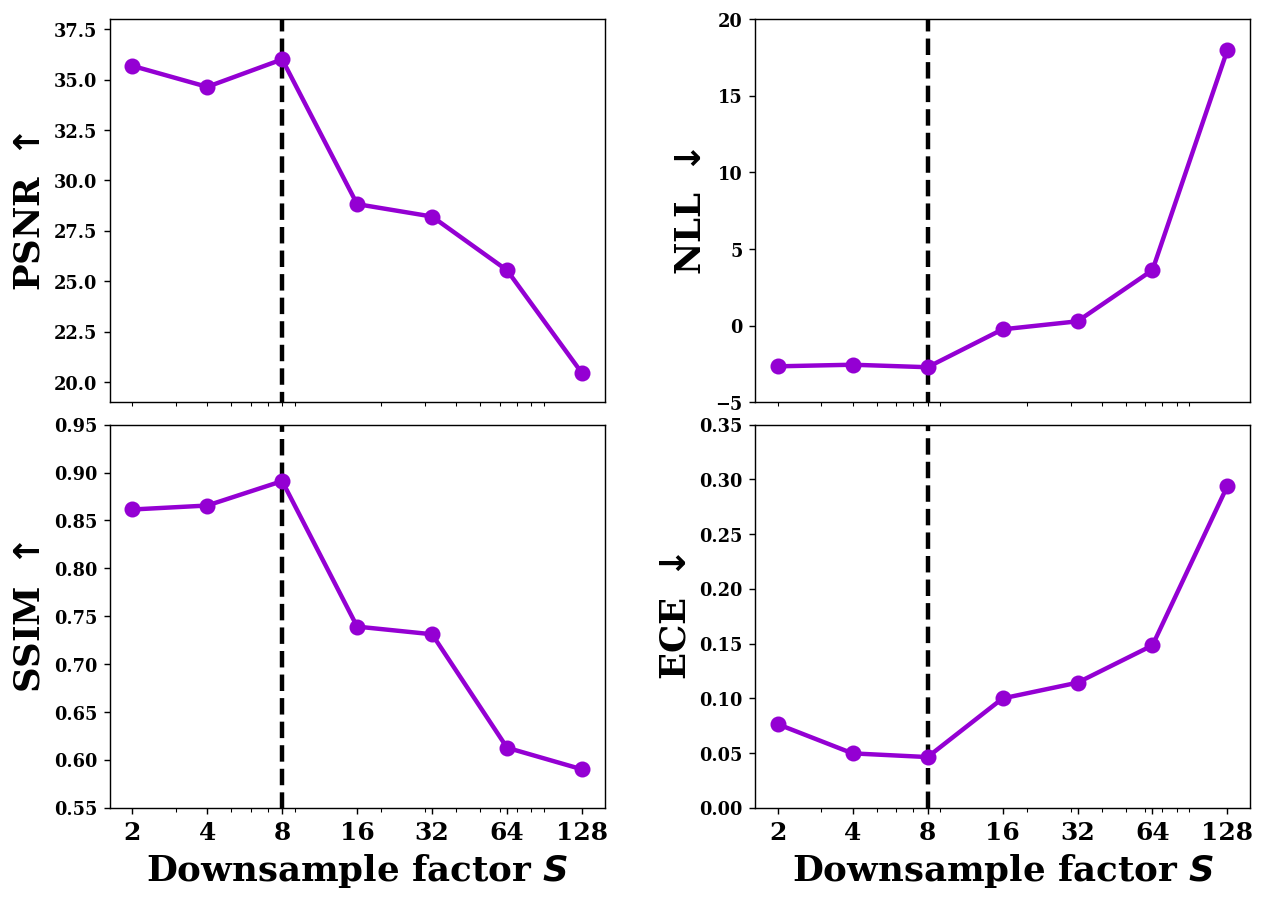}
    \caption{
        {\bf Ablation study on inherent dimensionality of ShuffleFlow.}
        We show an optimal downsample factor of 8 (marked by a vertical dashed line) for both image quality and uncertainty calibration.
    }
    \label{fig:ablation_dim}
\end{figure*}

As shown in Fig.~\ref{fig:ablation_dim}, we explore the impact of the dimensionality of ShuffleFlow (purple) by adjusting the pixel-unshuffling downsampling factor $S$.
For image size of 256$\times$256, the ShuffleFlow dimensionality is $d=\frac{256^2}{S^2}$.
We observe an optimal performance in image quality and uncertainty calibration at 2$^{10}$, corresponding to a downsampling factor $S=8$.
The computational cost grows slowly with dimensionality and exhibits a sudden increase after 2$^{12}$. 
This is because most of the computation is dominated by the score prior when $d \leq 2^{12}$, while the flow network dominates when $d > 2^{12}$.
This experiment is conducted on compressed-sensing MRI with a score prior.

\begin{figure*}
    \centering
    \includegraphics[width=0.9\linewidth]{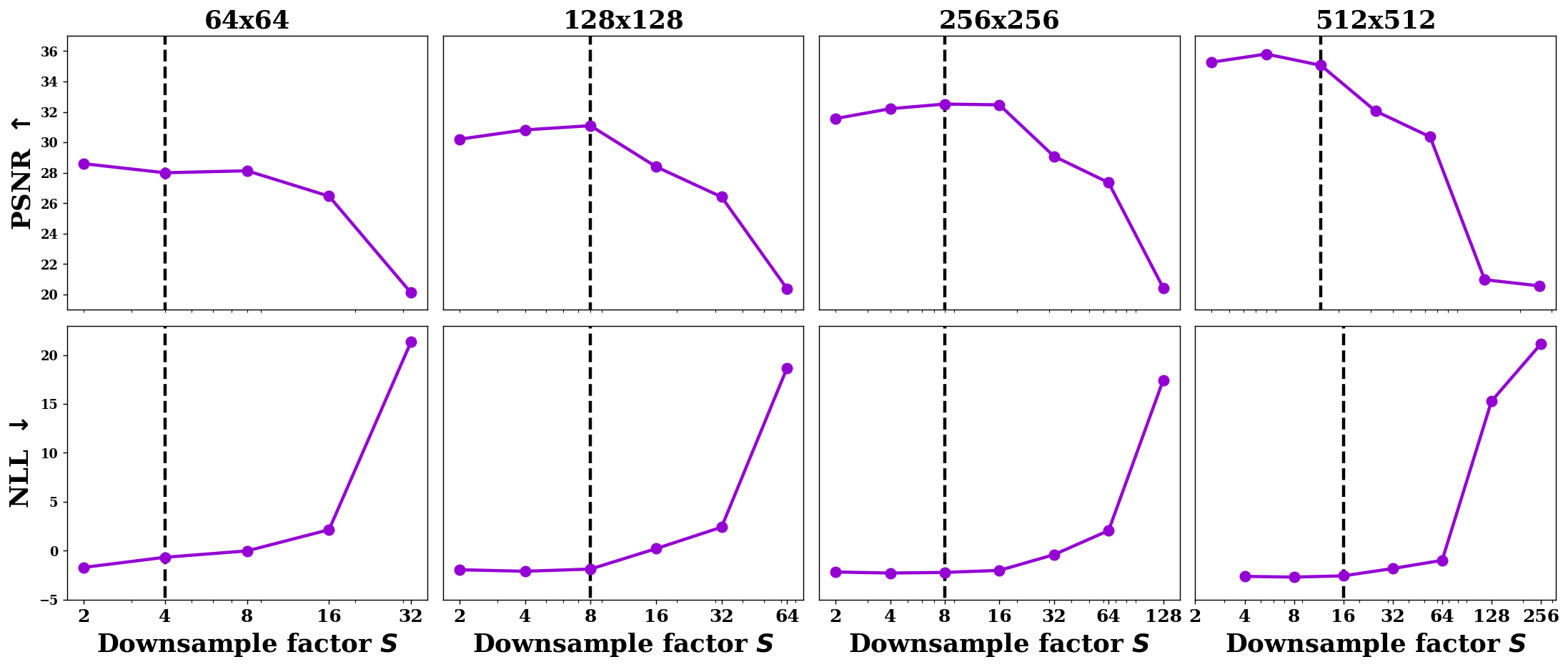}
    \caption{
        {\bf Ablation study on inherent dimensionality of ShuffleFlow for different image sizes.}
        We show downsampling by $S=\sqrt{N}/2$ (vertical dashed lines) preserves image reconstruction performance, while setting $S$ too large degrades images.
    }
    \label{fig:ablation_downsample}
\end{figure*}

We also explored the impact of the downsampling factor $S$ on several different image sizes (see Fig.~\ref{fig:ablation_downsample}).
For each image size, performance sharply decreases once $S$ becomes too large. In all observed cases, this happens after $\sqrt{N}/2$, allowing $O(N^2)$ scaling without a dramatic difference in image reconstruction performance.
This provides extended evidence that a downsampling factor of $S=\sqrt{N}/2$ (marked with vertical dashed lines) preserves performance, yielding an effective scaling of $O(N^2)$ in Eq.~\ref{eq:model_size}.
This experiment is conducted on compressed-sensing MRI with a TV prior, due to the expense of training new diffusion models for each image size, but note that it shows agreement with diffusion results.
We leave further validation of such observations to future work.

\stepcounter{fi}

\stepcounter{fi}

\subsection{Weights in KL divergence loss}

\begin{figure*}
    \centering
    \includegraphics[width=\columnwidth]{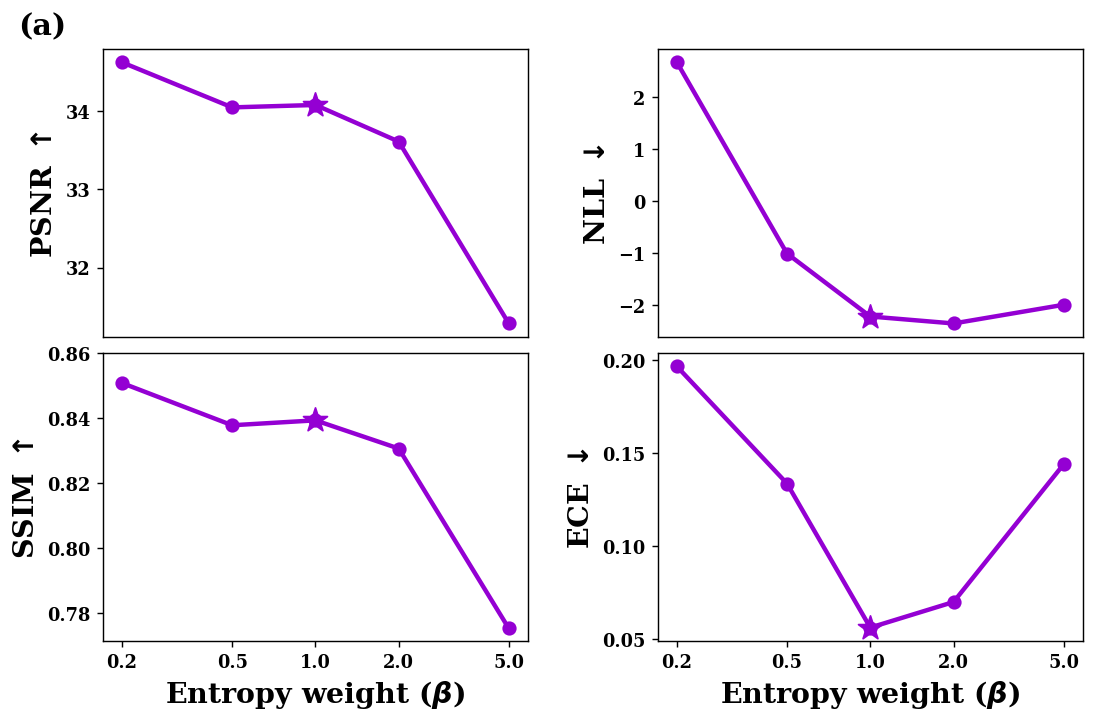}
    \includegraphics[width=\columnwidth]{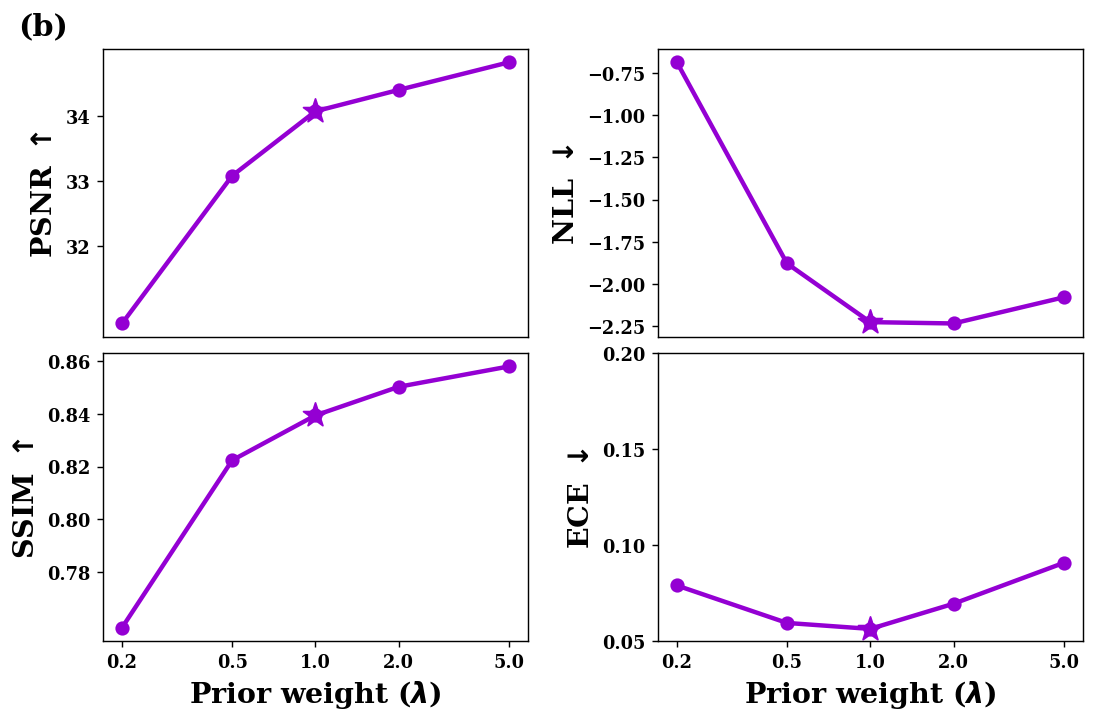}
    \caption{
        {\bf Ablation studies on VI loss weights.}
        (a) With the increase of entropy weight $\beta$, we see a decrease in image quality due to increased diversity in posterior samples.
        We select $\beta=1.0$ for good image quality and optimal uncertainty calibration (marked with a star).
        (b) With the increase of prior weight $\lambda$, we see an increase in image quality due to the stronger regularization effect of the score prior.
        However, the uncertainty also decreases due to the stronger constraint of the prior, resulting in optimal uncertainty calibration at $\lambda=1.0$ (marked with a star).
    }
    \label{fig:ablation_weights}
\end{figure*}

\begin{figure*}
    \centering
    \includegraphics[width=0.8\linewidth]{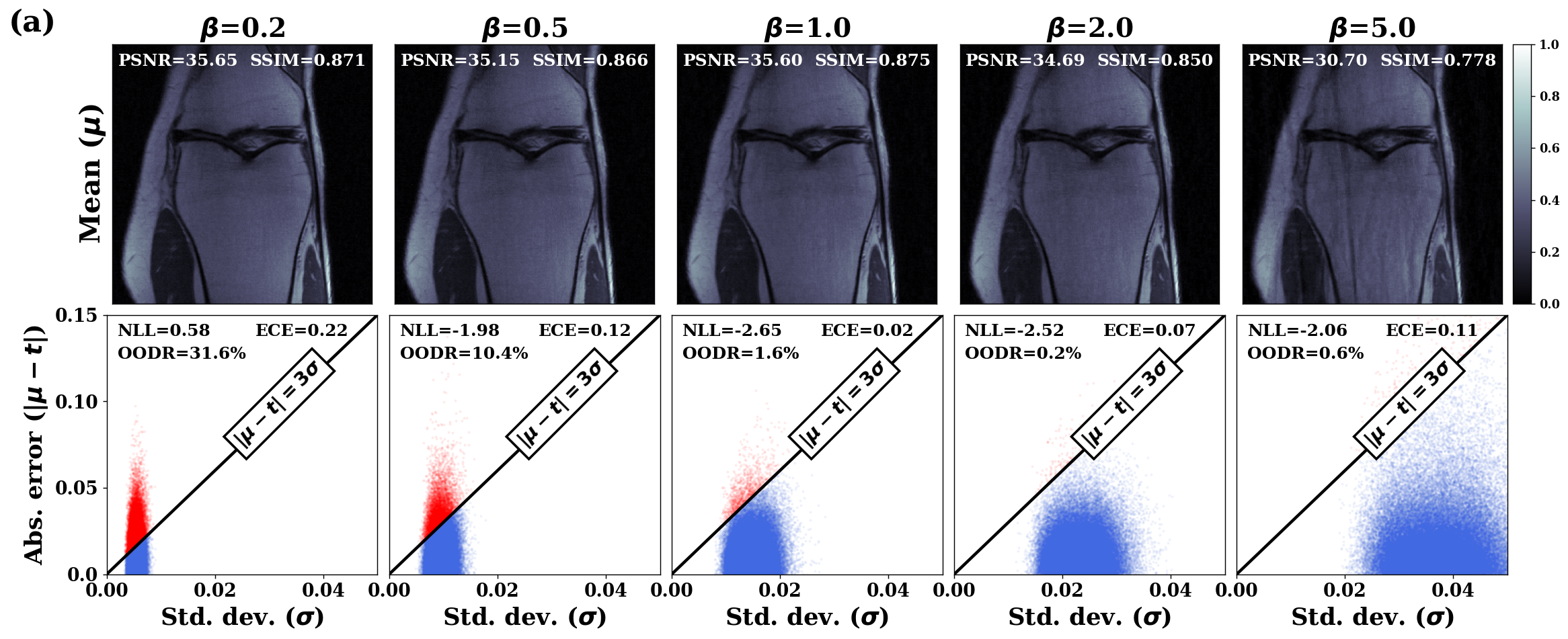}
    \includegraphics[width=0.8\linewidth]{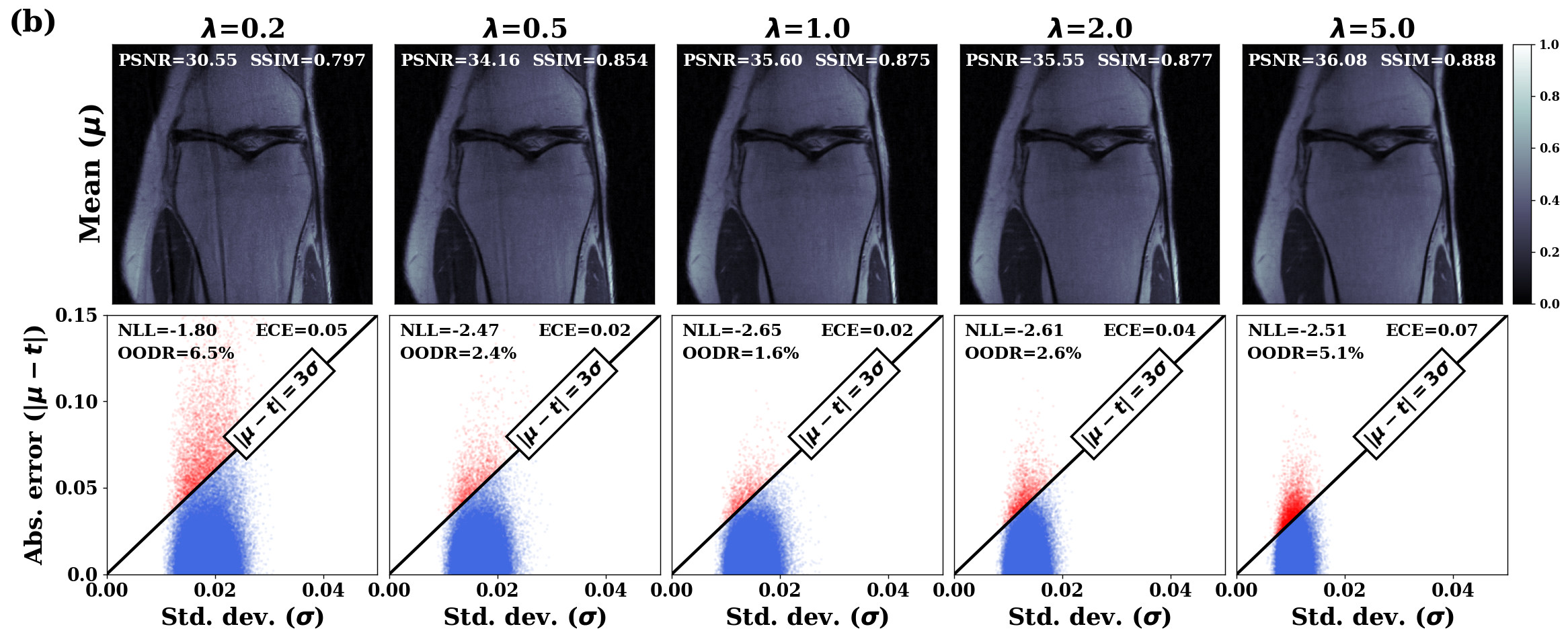}
    \caption{
        {\bf Visual samples of ablation studies on VI loss weights.}
        (a) With the increase of entropy weight $\beta$, we see a decrease in image quality (row 1) and an increase in uncertainty (see distribution shift towards the right in row 2).
        (b) With the increase of score prior weight $\lambda$, we see an increase in image quality (row 1) and a decrease in uncertainty (see distribution shift towards the left in row 2).
    }
    \label{fig:ablation_weights_sample}
\end{figure*}

We also explore the impact of entropy weight $\beta$ and score prior weight $\lambda$ in Eq.~\ref{eq:optimization} on compressed sensing MRI. 
Results are shown in Figs.~\ref{fig:ablation_weights} \& \ref{fig:ablation_weights_sample}.
We show that with the increase of $\beta$, the image quality decreases and the predicted pixel uncertainty also increases (see pixel distribution shift in row 2 of Fig.~\ref{fig:ablation_weights_sample}a).
Although the image quality is better with a small $\beta$, the uncertainty is underestimated, which leads to poorly calibrated uncertainty.
With a large $\beta$, our method predicts high uncertainty but also suffers from high pixel errors due to decreased image quality when the posterior samples are more diverse.
For the best calibrated uncertainty, we choose $\beta=1.0$ (marked with a star) for all flow-based methods in this paper.

With the increase of $\lambda$, we see an increase in image quality and a decrease in pixel uncertainty (see pixel distribution shift in row 2 of Fig.~\ref{fig:ablation_weights_sample}b).
This is because the score prior has a stronger regularization effect with larger $\lambda$, which increases image quality but also acts as a stronger constraint on the posterior distribution.
For the best calibrated uncertainty, we choose $\lambda=1.0$ (marked with a star) for all flow-based methods in this paper.

\stepcounter{fi}

\stepcounter{fi}

\subsection{Flow network architecture}

\begin{figure*}
    \centering
    \includegraphics[width=0.75\linewidth]{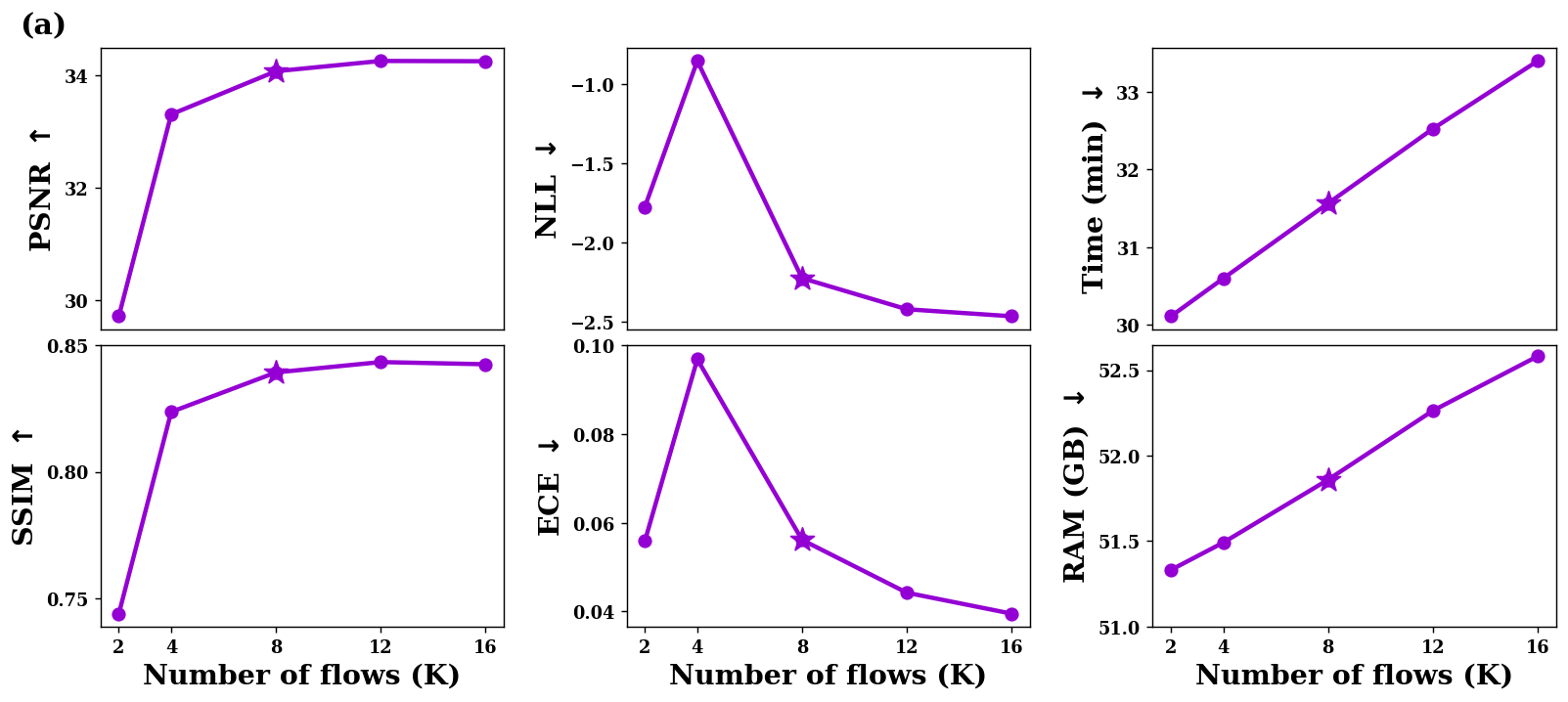}
    \includegraphics[width=0.5\linewidth]{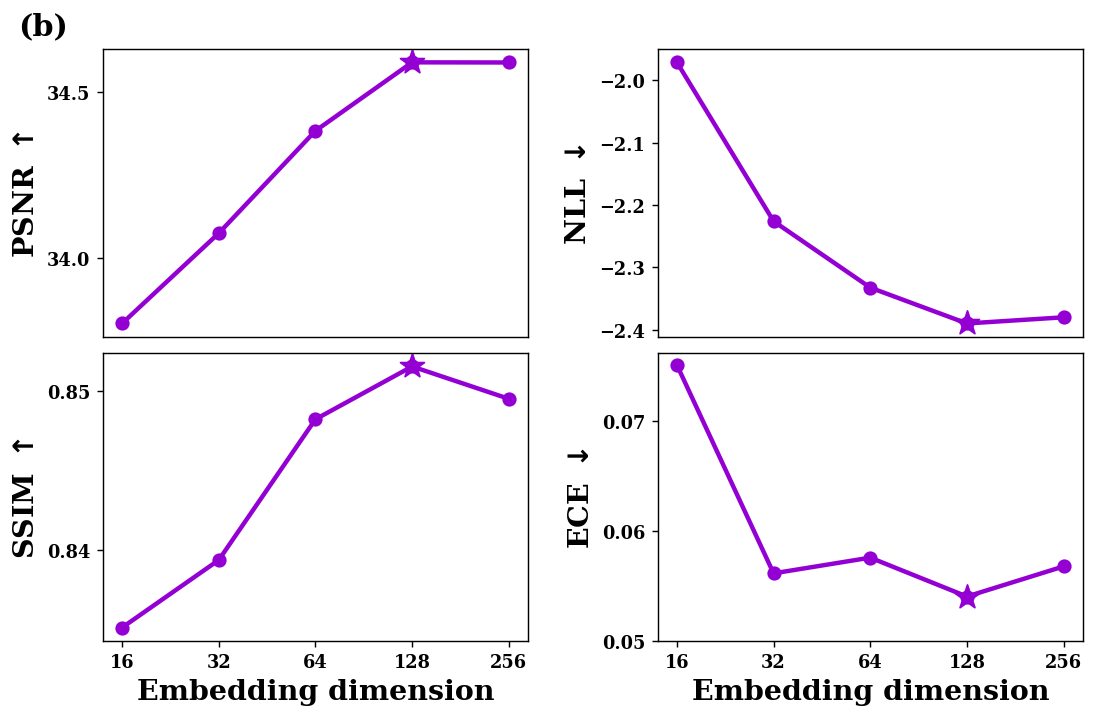}
    \caption{
        {\bf Ablation studies on network architecture.}
        (a) With the increase in the number of flow layers, we show an improvement in image quality, uncertainty calibration, and an increase in computational cost (marked with a star).
        We use 8 flow layers (marked with a star) in this performance vs. resource tradeoff.
        (b) We show a small increase in performance with the increase of the dimension of the embedding vector $\mathbf{h}_i$.
        We set the embedding vector dimension to 128 for optimal performance (marked with a star).
    }
    \label{fig:ablation_network}
\end{figure*}

We run ablation studies on the number of flow layers and the dimension of embedding vectors $\mathbf{h}_i$ in ShuffleFlow.
In Fig.~\ref{fig:ablation_network}a, we observe that more flow layers lead to better image quality and uncertainty, but exhibit a linear growth in optimization time and memory.
We choose 8 layers (marked with a star) in this performance-resource tradeoff space.
In Fig.~\ref{fig:ablation_network}b, we also see a slight increase in performance with a larger embedding dimension, which saturates around 128 (marked with a star).
We use 128-length embeddings for ours, ours w/ patch, and CF-NeRF.
Both experiments are done with compressed-sensing MRI with a score prior.

\begin{figure*}[b!]
    \centering
    \includegraphics[width=0.6\linewidth]{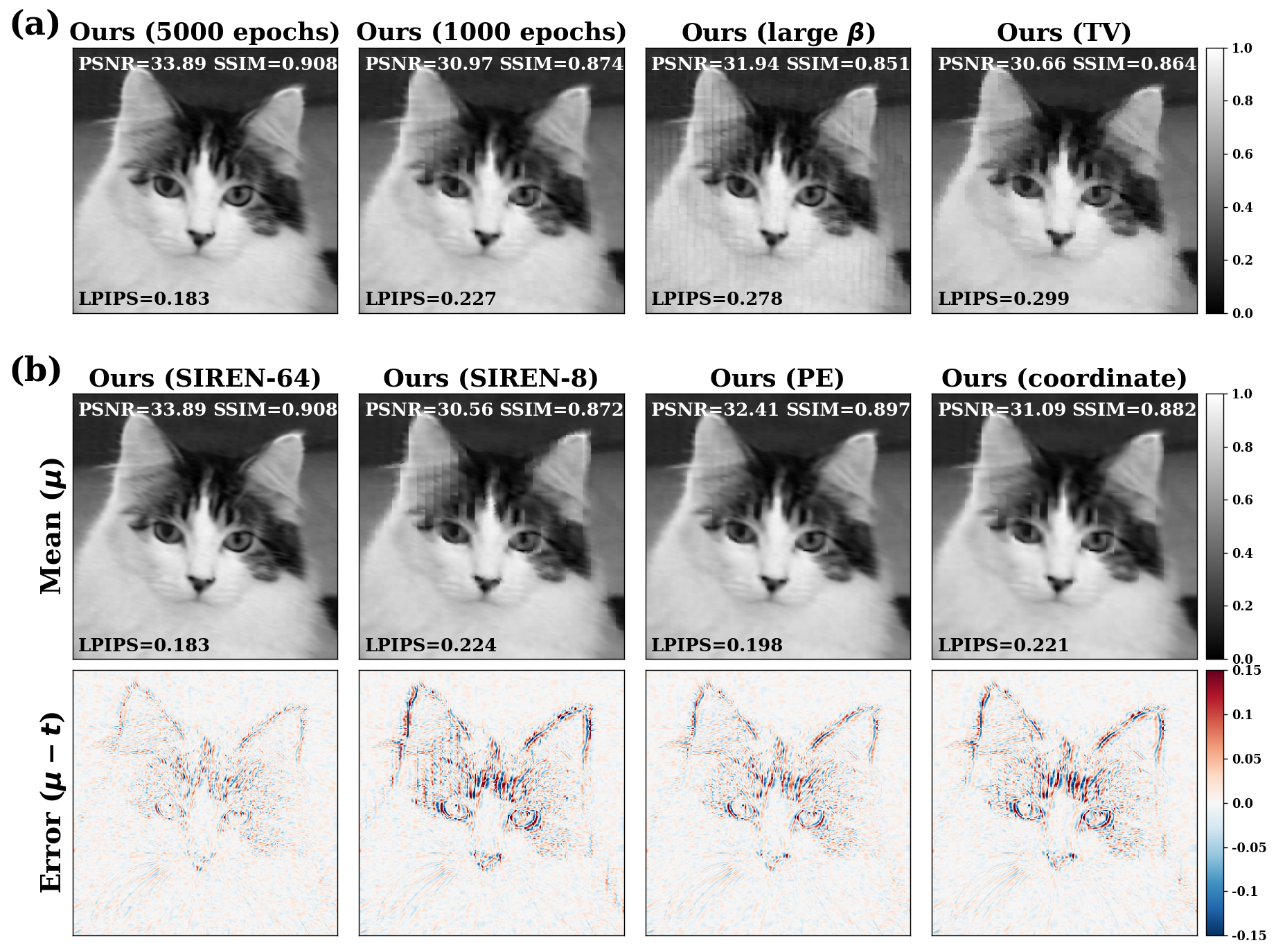}
    \caption{
        {\bf Additional ablation studies on patch-like artifacts and neural field.}
        (a) Failure cases with patch-like artifacts.
        We observe that non-ideal optimization settings lead to patch-like artifacts.
        (b) Ablations on the neural field encoder. We show that using a smaller neural field (SIREN-8), using positional encoding in place of the neural field (PE), and directly conditioning on the coordinates (coordinate) lead to decreased image quality.
    }
    \label{fig:ablation_nf}
\end{figure*}

In Fig.~\ref{fig:ablation_nf}a, we show that using non-ideal optimization settings (fewer training epochs, large entropy weight $\beta$, and using a TV prior instead of a score prior) leads to patch-like artifacts.
In Fig.~\ref{fig:ablation_nf}b, we show that using a smaller neural field, using positional encoding, and conditioning directly on the coordinates leads to decreased image quality.
These experiments are conducted on motion deblurring with a score prior.

\stepcounter{fi}

\stepcounter{fi}

\end{document}